\documentclass[]{elsart}
\date{ }    

\usepackage{graphicx}

\pdfoutput=1 

\newcommand{\shrink}[1]{} 

\begin{document}
\begin{frontmatter}
\title{Neural network learning of optimal Kalman prediction and control}
\author{Ralph Linsker}
\address{IBM T.J. Watson Research Center, Yorktown Heights, NY
10598, USA \\
Email: linsker@us.ibm.com
} 

\begin{abstract}

Although there are many neural network (NN) algorithms for prediction and for 
control, and although methods for optimal estimation 
(including filtering and prediction) and for optimal control in
linear systems were provided by Kalman in 1960 (with nonlinear extensions
since then), there has been, to my knowledge, no
NN algorithm that learns either Kalman prediction or
Kalman control (apart from the special case of stationary control).
Here we show how optimal Kalman prediction and control (KPC),
as well as system identification, can be 
learned and executed by a recurrent neural network composed 
of linear-response nodes, 
using as input only a stream of noisy measurement data.

The requirements of KPC 
appear  
to impose significant constraints on the allowed 
NN circuitry and signal flows.  The NN architecture implied by these
constraints bears certain resemblances to the local-circuit 
architecture of mammalian cerebral cortex.  We discuss these
resemblances, as well as caveats that limit our current ability to draw
inferences for biological function.  
It has been suggested that the local cortical circuit (LCC)
architecture may perform core functions 
(as yet unknown) that underlie sensory, motor, and other cortical 
processing. 
It is reasonable to conjecture that such functions may include 
prediction, the estimation or 
inference of missing or noisy sensory data, 
and the goal-driven generation of 
control signals.  
The resemblances found between the KPC NN 
architecture and that of the LCC 
are consistent with this conjecture.
\end{abstract}


\end{frontmatter}

\newpage

\section{Introduction}

The Kalman optimal filter and 
controller (Kalman, 1960) 
are classical solutions for efficient
optimal estimation (which includes
prediction, filtering, and smoothing)
and optimal control
in linear systems.
They also form the basis for extensions 
that yield approximately optimal solutions for 
certain types of nonlinear systems.
Within the field of neural networks, a great many
algorithms for prediction and control 
in a variety of settings 
have been developed.  Yet there exists, to my 
knowledge, no neural 
algorithm for learning the optimal Kalman filter (KF),
nor for learning the optimal Kalman controller (KC) except 
in the stationary case (discussed in section 5.2). 

In addition, the classical Kalman algorithms assume that the
parameters characterizing the external system (the `plant')
and the measurement process are known in advance.  When
they are not known, a separate process of system 
identification is typically performed.

In this paper we derive a neural network (NN) circuit and 
algorithm that learns and executes Kalman estimation and control, 
and that also
determines the required 
combinations of  
plant and measurement process parameters, 
using as input {\it only} a stream of
noisy measurement data vectors (and, for the controller, the 
specification of the cost function whose value is
to be optimized).
The 
differences between the
Kalman filter and controller learned by the network,
and those derived using the classical Kalman
algorithms, 
can be made arbitrarily small,
provided that certain expectation values over distributions are sufficiently 
well approximated by the corresponding finite-sample 
statistics\footnote{We use `sample' in its statistical sense, to mean 
a subset of a population (ensemble) that is defined by a distribution.}
(as discussed in Sections 3 and 4).

The resulting artificial neural circuit and algorithm may prove
useful for implementing the learning and execution of Kalman
prediction and control,
and its nonlinear extensions,
in parallel systems consisting of simple processors.

The resulting circuit architecture also has distinctive features 
that invite comparison with aspects of 
biological neural networks, particularly in cerebral cortex,
and may help in exploring the possible functions of
such networks.  

The paper is organized as follows.
Section 2 summarizes the optimal linear estimation 
and control problems, 
and the classical Kalman filter and controller
algorithms.
In section 3 we 
derive a neural network algorithm that 
both solves the system identification problem -- 
i.e., learns the dynamical properties of the plant 
(as these properties are reflected in the 
measurement data)
-- and learns 
the optimal Kalman filter with arbitrary accuracy.
The derivation proceeds in several stages.  We
first transform the classical Kalman 
estimation equations into
a form that explicitly involves only the input measurement
vectors, and not the plant  
state vectors themselves.  
(We do this because the plant state is unknowable from the data
in principle, even apart from noise,
when, as we assume, the transformation from plant state to 
measurement vector is not specified to the NN.)
We then derive a learning procedure 
that exactly implements 
these transformed equations, but
that is expressed in terms of certain expectation values.
Next, we derive a neural network algorithm that 
implements this learning procedure with arbitrary 
accuracy (depending upon how well the expectation values
are approximated by sample statistics).
Some distinctive elements of this algorithm
include: (a) the use of local neural network methods to
perform the required learning or use of the inverse of an
error covariance matrix; (b) the generation of this
covariance matrix by using either an ensemble of 
measurement vectors at each time step (e.g., the positions
of a set of tracked features in a visual scene), 
a single vector tracked over time to generate such an 
ensemble, or a combination of the two;  
(c) the simultaneous
learning of the Kalman filter,
use of that filter to predict the future plant state, 
and refinement of the learning of the plant dynamical parameters; 
and
(d) a specific recurrent circuit architecture,
and sequencing of computations, that
are implied by the algorithm.    
Finally, the joint NN learning of the plant dynamics and
the Kalman filter is illustrated with a numerical example.

In section 4, we derive a neural algorithm for Kalman
control.
The several stages of the derivation are 
similar to those used for Kalman
estimation.  There are evident
similarities that result from the mathematical duality (Kalman, 1960) 
between Kalman's optimal estimation and optimal control solutions,
but there is an additional distinctive feature:  
Kalman's duality includes a time reversal operation, so that 
the Kalman control matrix is computed 
by a process that operates `backward in time,' from the
future time of target (goal) acquisition to the present.
We show how this requirement is implemented within the
neural algorithm, which handles a decrementing time index
during the learning process, and 
generates a sequence of controller outputs to the plant 
in physical (forward-moving) time.
We then integrate the control method into the same
neural circuit and algorithm that handles estimation and
system identification. 

In section 5 we discuss several issues.  First, we identify
ways in which the  
computational task -- Kalman prediction and control -- 
places constraints on the type of NN circuitry and signal flows that
are involved in performing that task.  
Second, we comment on applications to artificial NN designs, and 
discuss prior work that 
has used NNs in conjunction with Kalman methods.

Finally, in subsection 5.3 and the speculative subsection 5.4,
we identify certain resemblances between 
the artificial NN that we are led to by the 
Kalman prediction and control (KPC) constraints,
and the architecture (and proposed signal flows) of the 
putative
`local cortical circuit' (LCC, minicolumn, canonical
microcircuit) of
mammalian cerebral cortex.
The resemblances between the KPC NN and the LCC, 
and important caveats that apply to the 
interpretation of these resemblances, are discussed.

Section 6 summarizes and concludes the paper.

\section{Classical Kalman linear estimation and control} 

In classical linear estimation and control (Kalman, 1960) 
an external system 
(the `plant') is described by a 
state vector $x_t$ (e.g., a point's trajectory) at each discrete 
time $t$, and
the dynamical rule 
\begin{equation}
x_{t+1} = F x_t + B u_t + m_t ~,
\label{eq:xevol}
\end{equation}
where $m_t$ is  
plant noise (e.g., random buffeting of an object) 
having zero mean and covariance $Q$,
and the optional vector $u_t$ is 
an external driving term and/or a computed control term.
Each measurement vector $y_t$ satisfies
\begin{equation}
y_t = H x_t + n_t ~,
\label{eq:yofx}
\end{equation}
where $n_t$ is 
measurement noise having zero mean and covariance $R$.
The matrices $F$, $B$, $H$, $Q$, and $R$, and the vector  
$u_t$, are assumed known. 
(Continuous-time versions of these problems 
and their Kalman solutions have been formulated,
but we will limit ourselves to the discrete-time case for simplicity.)

\subsection{Classical Kalman estimation (filtering and prediction)}

Given measurements through time $t$,	
the goal of optimal 
filtering (or, respectively, one-step-ahead prediction)
is to compute a posterior state estimate
$\hat{x}_{t}$ (resp., a prior state estimate $\hat{x}^-_{t+1}$)
that minimizes 
the generalized mean-square estimation error\footnote{Notation:
$E[\ldots]$ denotes expectation value, prime denotes transpose, 
and
$I$ is the identity matrix.}  
$E[(\xi_t)' C \xi_t]$ (resp., $E[(\xi^-_{t+1})' C \xi^-_{t+1}]$)
where
$\xi_t \equiv x_t - \hat{x}_t$,
$\xi^-_{t+1} \equiv x_{t+1} - \hat{x}^-_{t+1}$,
and
$C$ is a symmetric positive-definite matrix.
Throughout this paper, a variable having a `hat' will generally denote
an estimate of the underlying variable, and a variable having a tilde 
will denote the result of applying a transformation to the underlying
variable.   

Kalman (1960) 
showed that, under a variety of conditions,
the optimal estimation solution for both filtering and prediction
is given by what we will refer to as the `execution' equations 
\begin{equation}
\hat{x}_t = \hat{x}^-_t + K_t (y_t - H \hat{x}^-_t ) ~; \; \; \;
\hat{x}^-_{t+1} = F \hat{x}_t + B u_t ~;
\label{eq:xhat}
\end{equation}
and the `learning' equations 
\begin{equation}
K_t = P^-_t H' ( H P^-_t H' + R )^{-1} ~; \; \; \;
P^-_{t+1} = F (I -K_t H) P^-_t F' + Q ~.
\label{eq:KPm}
\end{equation} 
(These solutions are independent of $C$.)
Equations \ref{eq:KPm} are initialized by assuming some 
distribution of values for $\xi^-_0$
and setting $P^-_0 \equiv E[\xi^-_0 (\xi^-_0)']$.
It then follows (Kalman, 1960) 
that, for all $t$, $P^-_t = E[\xi^-_t (\xi^-_t)']$.
Thus the KF matrix, $K_t$, is learned iteratively
using Eqs.~\ref{eq:KPm}, starting 
with an arbitrary matrix and converging exponentially rapidly 
to its final value as each new 
measurement is obtained.  
The classical KF learning algorithm involves multiplications of one 
matrix by another, and matrix inversion.

The model prediction $\hat{x}^-_t$ and the current measurement
$y_t$ are optimally blended 
(to minimize the estimation error) by
using the KF (Eq.~\ref{eq:xhat}). 
As expected intuitively,
when the plant noise is much greater than the measurement noise,
this blending gives greater weight to the current measurement; 
when the measurement noise is much greater, the model prediction 
receives greater weight. 

\subsection{Classical Kalman control}
The classical control problem known as 
`linear quadratic regulation' 
can be defined as follows.
A controller is required to generate a set of signals 
$\{ u_t \}$ that 
minimizes the expected total cost $J$ of approaching a 
desired target state 
at time $N$.
Here $J$ reflects the cost of producing each control output 
(e.g., the energetic cost of moving a limb or
firing a rocket thruster) plus a penalty that is a function 
of the 
difference between the actual state at each time step and the 
target state.
Specifically,
\begin{equation}
J = E [ \Sigma_{t=t_0}^{N-1} (u'_t g u_t + x'_t r x_t) + x'_N r x_N ] ~~,
\label{eq:J}
\end{equation}
where
$g$ and $r$ are specified symmetric positive-definite matrices.
(We take the target state to be $x=0$ for simplicity.)

The classical Kalman control (KC) algorithm (Kalman, 1960), 
starting at a current time $t_0$,
computes during the learning step a sequence of matrices 
$\{ L_N, S_{N-1},$ $L_{N-1},$ $S_{N-2}, \ldots ,$ 
$S_{t_0 +1}, L_{t_0} \}$, 
where each $L$ is a KC matrix and each $S$ is an auxiliary 
symmetric positive-definite matrix, with
$S_N = r$.  The matrices are iteratively computed using
\begin{equation}
L_{\tau} = (B' S_{\tau} B + g)^{-1} B' S_{\tau} F ~~; 
S_{\tau -1} = (F' - L'_{\tau} B') S_{\tau} F + r ~~.
\label{eq:LS}
\end{equation}
Once these matrices have been computed, 
the time $t$ is incremented starting from $t_0$, and
the execution step 
$u_t = - L_t \hat{x}_t$
for each $t = t_0, t_0+1, \ldots, N-1$
generates the optimal control signals $\{ u_t \}$.

Thus the Kalman controller generates an optimal sequence 
of control outputs that 
cause the plant to approach a desired target state at a 
specified future time $N$. 
The KC matrices are learned iteratively, 
`backward in time,' then are executed 
forward in time. 

For both Kalman prediction and control, several of the parameters
that we have taken above to be constant in time -- e.g., $F$, $H$,
$Q$, $R$, $g$, and $r$ -- may in fact vary slowly compared with the
time scale over which KPC learning occurs, or may change abruptly to
new values.  In these cases, KPC can still yield approximately
optimal results (after a transitional period of adjustment, in the 
case of an abrupt change).  The same is true for the neural 
KPC algorithms we derive below.  Even when these parameters are allowed
to vary with time, however, we will suppress
the time index for simplicity.          

\section{Neural Kalman estimation and system identification}

\subsection{Transformation from plant variables to measurement variables}

For our neural algorithm 
(in contrast to Kalman's solution above), 
we assume that the NN is given {\it only} the stream of noisy 
measurements $\{ y_t \}$;
no plant, measurement, or noise covariance parameters are 
assumed known. 
Since the transformation $H$ from the plant state to the 
measurement vector 
is not specified to the network, the plant state $x$ is unknowable
in principle, 
and we therefore work in the space of the measurement vector $y$.
	 
We eliminate $x$ in favor of $y$ variables as follows. 
The goal of classical Kalman estimation
is to produce estimates $\hat{x}$ (or $\hat{x}^-$)
that best approximate the true plant state $x$. 
This is equivalent to producing estimates $\hat{y}=H\hat{x}$ 
(or $\hat{y}^- = H\hat{x}^-$)
that best approximate $Hx$.  We  
define $Y_t \equiv Hx_t$, which we call
the `ideal noiseless measurement' of $x_t$.
The goal of optimal filtering (respectively, prediction) stated earlier
is then, given $\{ y_0, \ldots, y_t \}$,
to compute a posterior (resp., prior) measurement estimate $\hat{y}_t$ 
(resp., $\hat{y}^-_{t+1}$)
that minimizes 
$E( \zeta'_t \tilde{C} \zeta_t )$ where
$\zeta_t \equiv Y_t - \hat{y}_t$ (resp.,
$\zeta_t \equiv Y_{t+1} - \hat{y}^-_{t+1}$),
and
where $\tilde{C} \equiv H'CH$. 
(As in the classical derivation, the resulting optimal KF
is independent of $\tilde{C}$.)

The basic outline of the algorithm is as follows.
We do not need to learn $H$ or $F$ 
-- and indeed they are not individually knowable 
by the 
network\footnote{We therefore use the term `system identification' to mean 
the determination of the plant dynamics as these dynamics are reflected
in the measured quantities $y$ -- e.g., the determination of 
the matrix $\tilde{F}$ rather than $F$.}; we need only to learn
the combination\footnote{$M^+$ denotes the 
Moore-Penrose pseudoinverse of $M$.
When $M$ is a square matrix of full rank, $M^+ \equiv M^{-1}$.} 
$\tilde{F} \equiv HFH^+$.
If we had access to the ideal noiseless measurements $Y$ --
which we do not -- we could
learn $\tilde{F}$ by minimizing the mean-square prediction error,
$E[ (Y^{\rm pred}_t - Y_t )' (Y^{\rm pred}_t - Y_t ) ]$, with respect
to $\tilde{F}$,
where $Y^{\rm pred}_t = \tilde{F} Y_{t-1} + \tilde{u}_{t-1}$   
and $\tilde{u}_t \equiv HBu_t$.  
In practice, 
we use two quantities as surrogates for the unknown $Y$, each at a different 
stage of the algorithm, as follows.

At the start, neither $\tilde{F}$ nor the KF is known to the 
algorithm.  
First, $\tilde{F}$ is learned approximately, using 
the raw (noisy) measurement vector $y_t$ as a surrogate for $Y_t$, 
and the prediction
$y^{\rm pred}_t = \tilde{F}_{t-1} y_{t-1} + \tilde{u}_{t-1}$
as a surrogate for $Y^{\rm pred}_t$.  Here $\tilde{F}_{t-1}$ is the 
approximation, as computed at step $t-1$, to the true $\tilde{F}$.
We thus want to perform gradient descent on  
$E( \epsilon'_t \epsilon_t )$ with respect to $\tilde{F}_{t-1}$, where
$\epsilon_t = \tilde{F}_{t-1} y_{t-1} + \tilde{u}_{t-1} - y_t$. 
This yields the learning rule:
\begin{equation}
\tilde{F}_t = 
\tilde{F}_{t-1} - \gamma_F \, \partial E(\epsilon'_t \epsilon_t ) / 
\partial \tilde{F}_{t-1}
= \tilde{F}_{t-1} - \gamma_F E( \epsilon_{t} y'_{t-1} ) ~~,
\label{eq:Ftil0}
\end{equation}
where $\gamma_F$ is a learning rate.

Second, we start to learn the KF.
KF learning proceeds by learning a matrix $Z$ (defined below),
which is uniquely related to the classical KF matrix $K$.
We start to learn $Z$ (Eq.~\ref{eq:Zevol} below), 
using the current value of the learned $\tilde{F}$.
We (optionally) continue to refine $\tilde{F}$, still using Eq.~\ref{eq:Ftil0}, 
so that $Z$ and $\tilde{F}$ learning proceed in tandem.

Third, 
at some point $Z$ will have
been learned to a sufficiently 
good approximation that the resulting
estimates $\hat{y}$ (using Eq.~\ref{eq:yhat} below)
are comparable or superior 
to the raw measurements $y$, 
as estimates of 
the ideal noiseless measurement $Y$. 
At this point, while we continue to refine the learning of $Z$, we can also
refine $\tilde{F}$ further by using 
$\hat{y}_{t-1}$ and $\eta_{t}$ 
(Eqs.~\ref{eq:yhat} and \ref{eq:eta} below)
in place of $y_{t-1}$ and $\epsilon_{t}$ respectively; 
i.e., replacing Eq.~\ref{eq:Ftil0} by
\begin{equation}
\tilde{F}_t = \tilde{F}_{t-1} - \gamma_F E( \eta_{t} \hat{y}'_{t-1} ) ~~. 
\label{eq:Ftil1}
\end{equation}

Note that if the plant or measurement process parameters 
are not constant, but change 
slowly in time (perhaps as the result of a weakly nonlinear plant or 
measurement process), 
then
learning of $\tilde{F}$ and $Z$ (and $R$, if 
the measurement process is changing)
should continue.
If parameters change abruptly, so that the already-learned 
$Z$ and $\tilde{F}$ 
are no longer valid, one should then return to the first stage of the 
algorithm above; i.e., learn $\tilde{F}$ using raw measurements, and
suspend $Z$ learning until the new $\tilde{F}$ has been
sufficiently well learned.
A significant increase in the distance between $\hat{y}$ and $y$
indicates that such an abrupt change may have occurred and that
relearning is needed.

We turn now to the learning of $R$ and $Z$.

The measurement noise covariance
$R = E(n_t n'_t )$ is learned from `measurements' that are taken in 
`offline sensor' mode, 
i.e., in the absence of external input, so that  
$y_t = n_t$.  This is customary in KF practice.

To transform the classical KF learning Eqs.~\ref{eq:KPm}
into a form that will be suitable for a neural algorithm, 
we define
$Z_t \equiv H P^-_t H' + R$. 
Then 
$I-HK_t = R Z_t^{-1}$.
The transformed matrix equation that corresponds exactly to Eqs.~\ref{eq:KPm} is:
\begin{equation}
Z_{t+1} = \tilde{F} (I-RZ_t^{-1}) R \tilde{F}' + HQH' + R
\label{eq:Zevol}
\end{equation}
(see Appendix A.1 for proof). 

The execution equations are (cf. Eqs.~\ref{eq:xhat})
\begin{equation}
\hat{y}_t = y_t + R Z_t^{-1} \eta_t~~; 
\; \; \;
\hat{y}^-_{t+1} = \tilde{F} \hat{y}_t + \tilde{u}_t 
\label{eq:yhat}
\end{equation}
where 
\begin{equation}
\eta_t \equiv \hat{y}^-_t - y_t ~~.
\label{eq:eta}
\end{equation}
The vector
$\eta_t$ evolves according to
\begin{equation}
\eta_{t+1} = -y_{t+1} +  \tilde{F} (y_t + R Z_t^{-1} \eta_t) + \tilde{u}_t ~~
\label{eq:etaevol}
\end{equation}
(by Eqs.~\ref{eq:yhat} and \ref{eq:eta}), 
and is initialized by assuming some distribution of
values for $\eta_0$. 

A crucial point:  We are going to use an {\it ensemble} of $\eta_t$ values
to update the matrix $Z_t$. 
Each $\eta_t$ value evolves, via Eq.~\ref{eq:etaevol},
from its initial value $\eta_0$, where the $\eta_0$ ensemble has a
specified distribution.
(Similarly, in the classical Kalman formulation, 
each $x_t$ value evolves from its initial value $x_0$, and the
$x_0$ distribution is specified.
The use of 
this distribution enables Kalman to define the expectation value that is to
be minimized; see the text above Eq.~\ref{eq:xhat}.) 
Later, in section 3.2, we will approximate ensemble averages by 
suitable finite-sample averages, to derive a practical NN algorithm.  

Accordingly, 
we derive the following relationship between $E(\eta_t \eta'_t)$ and
$Z_t$: 
If $Z_0 = E(\eta_0 \eta'_0)$,  
then
Eqs.~\ref{eq:Zevol} and \ref{eq:etaevol} yield 
\begin{equation}
Z_t = E(\eta_t \eta'_t) 
\label{eq:Zeta}
\end{equation}
for all $t>0$.
This is proved by mathematical induction in
Appendix A.1.  
Thus Eqs.~\ref{eq:etaevol} (applied 
to each $\eta$ of an ensemble) 
and \ref{eq:Zeta} can be used for learning $Z$,
in place of Eq.~\ref{eq:Zevol}.

Note that $Q$, the plant noise covariance matrix, does not appear 
explicitly in
Eq.~\ref{eq:etaevol}.  Thus, the fact that $Q$
is not specified to the network poses no difficulty here.
This is in contrast to the classical Kalman
algorithm, which does assume $Q$ to be specified, and in which $Q$
does explicitly appear (see Eq.~\ref{eq:KPm}). 
The effects of $Q$ are instead captured 
implicitly in Eq.~\ref{eq:etaevol} 
as part of the behavior of the measured time series $\{ y_t \}$.  

At this point, we have gone part of the way toward a neurally implementable
algorithm:  
\begin{enumerate}
\item Equation \ref{eq:etaevol} requires the multiplication of a vector  
by a matrix -- not the multiplication of a matrix by a matrix (in contrast to the
classical Eqs.~\ref{eq:KPm}).  In a NN of the type we are 
considering here, a vector is 
represented
by the activities over a set of nodes, and a matrix by a set of connection 
strengths joining one set of nodes to the same or a different set 
of nodes.  Thus
matrix-times-vector multiplication is a natural operation for a NN, while 
matrix-times-matrix multiplication is not.  
\item Each of the learning rules -- for $\tilde{F}$, $R$, and $Z$ -- involves the
computation of a time-varying expectation value over a product of activity vectors.
We will approximate each expectation value 
by a sample average, using one of the methods described in 
section 3.2 below.
\item Equation \ref{eq:Zeta} yields a simple learning rule for $Z$, 
but we must compute the product $Z_t^{-1} \eta_t$,
involving the matrix inverse of $Z$, in 
Eq.~\ref{eq:etaevol}.  In section 3.3 
we describe two ways to resolve this difficulty.  
\end{enumerate}

\subsection{Learning of expectation values}

To learn $\tilde{F}$, $Z$ (or $Z^{-1}$), and $R$, we need in each case to 
compute a sufficiently good approximation to an expectation value matrix, 
denoted generally by $M_t \equiv E(v_t z'_t)$ 
where $v_t$ and $z_t$ are activity vectors
drawn at time $t$ from distributions that are in general time-varying.
We will approximate each expectation value
by a sample average,
where the sample is obtained by one of three methods: 
(a) combining contributions from multiple instances
$\{ v_t(p), z_t(p) \}$ indexed by  
$p=1, 2, \ldots, N_{\rm feat}$
(e.g., multiple features of a scene at each time step);
(b) time-averaging over a set of recent time steps
(e.g., using a recency-weighted average);
or (c) both.  A learning rule for all three methods is  
\begin{equation}
M_{t+1} = (1-\gamma_M) M_t + \gamma_M \langle v_t(p) z'_t(p) \rangle_p ~~.
\label{eq:Mlearn}
\end{equation}
For case (a), we set $\gamma_M = 1$; 
for case (b), $N_{\rm feat} = 1$ and 
$0 < \gamma_M \ll 1$;
and for case (c), $N_{\rm feat}>1$ and $0< \gamma_M <1$.

The notation $\langle \ldots \rangle_p$ can denote, in accordance
with its customary meaning, a batch average
over features:
$\langle m_t(p) \rangle_p \equiv (1/N_{\rm feat}) 
\sum_{p=1}^{N_{\rm feat}} m_t(p)$.
Alternatively, we can use it as a shorthand to denote  
the incremental updating of $M$ using the contribution of the $N_{\rm feat}$ 
features, one
feature at a time.  (Such an incremental update will be used in the
numerical example of section 3.6.)  
 
The terms on the RHS of Eq.~\ref{eq:Mlearn} are, respectively, 
a `forgetting' term and one or more Hebbian learning terms.
For the learning to be Hebbian, 
the activity components in the product term $v^i z^j$ must be the
activities present at either end of the connection $M^{ji}$ that is updated by
that product term.\footnote{    
Some of the learning rules used in this paper will contain a term that
{\it decreases} a connection strength when the product of the two activities 
is positive.  This is sometimes referred to as `anti-Hebbian' learning; in this 
paper this distinction is unimportant, since trivial changes in the algorithm
can change the signs of activity terms without affecting the overall
computation.  Thus learning rules that modify $M_t$, by an amount proportional
to a product of the activities at the two ends of each connection, will
be called `Hebbian' regardless of sign.}        

Some caveats apply to each of these three methods.

Regarding method (a):  This method
would ideally sample one feature $p$ from each of multiple
independent external systems, each such system being 
governed by the same plant dynamics.
However, in practice, multiple features are sampled from a single system.  Use of 
method (a) thus assumes that $\langle v_t(p) z'_t(p) \rangle_p$ provides   
a sufficiently unbiased estimate of $E(v_t z'_t)$.

Regarding method (b):
Here the number of past
time steps being sampled, $T_{\rm samp}$, is of $O(1/\gamma_M)$.  
The slower the learning rate $\gamma_M$, the 
less noisy will be the stochastic gradient-descent update of the weight matrix,
but the more time steps will be required for convergence to the optimal KF.
For this method we require that $T_{\rm samp}$ 
should be large enough so that fluctuations in the 
recency-weighted time average are kept small, yet small enough so that
the true time-dependent expectation value $E(v_t z'_t)$ does not vary
substantially during the interval $T_{\rm samp}$.  

When the two conditions for method (b) cannot be jointly satisfied
[i.e., when 
$E(v_t z'_t)$ varies rapidly, so that 
$T_{\rm samp}$ must be kept small, and 
as a result the fluctuations are too large],
and/or when the available number of features
$N_{\rm feat}$ in method (a) is too small to keep fluctuations in check,
the combined method (c) may provide a workable solution.

Also note that when method (b) or (c) is used to learn $Z$ (or $Z^{-1}$), 
it will require $O(1/\gamma_Z)$ time steps of the neural algorithm 
to learn the changes in the KF matrix that occur in one time step 
using the classical Kalman algorithm.  

If one wants to implement a NN using method (a) or (c), 
with simultaneous processing of multiple features, 
and with batch updating of a single weight matrix 
that is then used for processing each of the features,  
one approach is to use `weight tying,' a standard 
NN technique in which a computed weight matrix is 
`copied' to multiple sets of connections (see, e.g.,
Becker \& Hinton, 1992).  
Alternatively, 
intermediate results of the computation for each $p$
can be sent to
a part of a hardware NN 
that processes 
each $p$ in turn.
The weight tying technique is  
not local, and the 
alternative technique involves circuitry beyond what we discuss here. 
Owing to its nonlocality, it is quite unclear how
or whether weight tying 
could be used in a biologically plausible network that is 
based on neural coding similar to that used in this paper.
(It might be more biologically 
plausible to accomplish the desired goal -- updating 
a transformation function 
and applying the same updated function 
to the processing of multiple features -- 
if one instead uses, e.g., population coding. 
This issue is outside the scope of the present paper.)
For software NN implementations, of course, weight tying 
poses no complication.  

Note that the fluctuations in the neurally computed KF vanish in the limit 
that the sample average 
over the set of $\eta \eta'$ values
converges to 
$E(\eta \eta')$.
This convergence is not theoretically
assured if the instances of 
$\eta$ in the sample 
have mutual dependencies; e.g., if multiple features
do not obey the same dynamical equations independently and with 
independent plant noise.
The extent to which convergence of the 
neurally computed KF to the classical KF 
may, in practice, be affected by such dependencies is an open question.

For the neural learning of Kalman {\it control}, on the other hand, the 
expectation values to be approximated are over a sample 
of internally
generated quantities (as discussed in Section 4). 
These quantities are independently drawn from the ensemble, and
we can choose as many instances of them as we wish
(subject to hardware or computational limits).  
As a result,
the possible dependencies among (and 
the limited number of) trackable features of the external plant
play no role in KC learning 
(except indirectly, through 
the learning of $\tilde{F}$).  

\subsection{Neural learning and use of the inverse of a covariance matrix} 

To implement Eqs.~\ref{eq:etaevol} and \ref{eq:Zeta}, we may either 
(1) represent and learn a quantity linear in $Z$
as a connection matrix, and use this matrix to
compute $Z^{-1}\eta$ from $\eta$, or 
(2) represent $Z^{-1}$ directly as a connection matrix, in 
which case we need a learning rule that updates $Z^{-1}$.  
We consider each option in turn.

Method 1 -- learning a quantity linear in $Z$: 
From Eq.~\ref{eq:Zeta} we obtain the update rule
\begin{equation}
Z_{t+1} = (1-\gamma_Z) Z_t + \gamma_Z \langle \eta_t \eta'_t \rangle_p ~~,
\label{eq:Zlearn}
\end{equation}
where 
$\gamma_Z$ is the learning rate.
We compute $Z^{-1}\eta$ as follows (Linsker, 1992): 
Use a matrix of lateral connections to represent $\tilde{Z}_t \equiv I-cZ_t$,
where $c$ is a scalar quantity.
$\tilde{Z}$ is learned using
\begin{equation}
\tilde{Z}_{t+1} =  (1-\gamma_Z) \tilde{Z}_t + \gamma_Z I - \gamma_Z c 
\langle \eta_t \eta'_t \rangle_p.
\label{eq:Dlearn}
\end{equation}
Then, suppressing the time index $t$, we have 
$(Z^{-1}) \eta = c (I-\tilde{Z})^{-1} \eta
= c [\eta + \tilde{Z}\eta + \tilde{Z}^2\eta + \ldots]$; this series converges
provided $c$ is chosen (Linsker, 1992) 
such that all eigenvalues of $\tilde{Z}$ lie within
the unit circle. 

Now, for the NN implementation, 
consider a set of nodes, equal in number to the dimension $D_{\rm y}$ of the 
vector $\eta$, and with the lateral connection from 
node $i$ to $j$ having strength $\tilde{Z}^{ji}$.
At a given time step $t$, perform the following sequence of steps
(using dynamics that are fast compared to the interval between $t$ and 
$t+1$, and are 
also fast compared to the `micro' time scale defined later):
Hold the feedforward input activity at node $i$ fixed and equal to 
$v^i_{\rm in} = \eta^i$.  
Prior to any passes through the lateral connections, the 
activity at node $i$ is thus $v^i(0) = \eta^i$.  
(In this paragraph, the argument of $v$ denotes the 
number of iterative passes
that have been made through the lateral connections.) 
Then, 
after $n$ passes (for each of 
$n = 1, 2, \ldots$), the activity
at node $j$ is $v^j(n) = \sum \tilde{Z}^{ji} v^i(n-1) 
+ v^j_{\rm in}$.
The asymptotic result is thus
$v^j(\infty) = (I-\tilde{Z})^{-1} \eta$, as required (apart from a final 
multiplication by $c$).
In practice (Linsker, 1992), 
several passes (at each value of the time index $t$) typically suffice to
approximate the asymptotic result.

Although it is the matrix $\tilde{Z}$ (rather than $Z$ itself)
that is implemented as a set of
connections using this method, we will for convenience refer to this 
method as learning a matrix of $Z$
connections, to distinguish it from $Z^{-1}$ learning below.  

Method 2 -- learning $Z^{-1}$: Initialize $(Z^{-1})_0$ to be an arbitrary symmetric
positive-definite matrix of lateral connections.  
Then, at each time step $t$ (and for each 
feature if there are multiple features),
compute $v_t = (Z^{-1})_t \eta_t$ by taking one pass through the lateral connections.
Use the learning rule that is derived in (Linsker, 2005): 
\begin{equation}
(Z^{-1})_{t+1} = (1+\gamma_Z) (Z^{-1})_t 
- \gamma_Z \langle v_t v'_t \rangle_p ~~.
\label{eq:Zinvlearn}
\end{equation}
This rule is valid
provided (a) that $\gamma_Z \ll 1$ and (b) that there is some mechanism either for
keeping $(Z^{-1})_t$ symmetric or for suppressing asymmetries in 
$(Z^{-1})_t$ that might arise from noise or roundoff error.
[For an artificial NN implementation in software, or in a 
hardware network having bidirectional connections, 
one can simply keep $Z^{-1}$ symmetric by fiat.  In a 
biological implementation, or in a NN having unidirectional connections in 
which $(Z^{-1})^{ij}$ need not equal $(Z^{-1})^{ji}$,
an asymmetry suppression mechanism would be needed to avoid instability.]
See (Linsker, 2005) 
for proof and details.  

Although the use of Method 2 imposes a special 
requirement (i.e., maintenance of $Z^{-1}$ symmetry), it avoids the 
need to iterate within a given time step $t$
that Method 1 entails.

\subsection{Neural network algorithm}

The resulting equations for learning and execution of Kalman 
estimation and system identification, and the sets of computations that carry 
out these processes, are as follows.  

(1) For early learning of $\tilde{F}$ (using raw measurement data):  
Eq.~\ref{eq:Ftil0}
yields 
\begin{equation}
\tilde{F}_t = \tilde{F}_{t-1} 
- \gamma_F \langle \epsilon_t y'_{t-1} \rangle_p ~~.
\label{eq:Ftil0learn}
\end{equation}

(2) For `offline sensor' learning of $R$ -- i.e., each `measurement' 
$y_t$ is a pure measurement
noise term, $y_t = n_t$: 
\begin{equation}
R_t = (1-\gamma_R) R_{t-1} + \gamma_R \langle n_t n'_t \rangle_p ~~.
\label{eq:Rlearn}
\end{equation}

(3) $Z$ or $Z^{-1}$ learning:  
For learning of $\tilde{Z} \equiv I-cZ$, use Eq.~\ref{eq:Dlearn}.   
Alternatively, for learning of $Z^{-1}$, use Eq.~\ref{eq:Zinvlearn}.

(4) For execution of Kalman estimation (i.e., the computation of 
optimal estimates using $Z$): 
The error activity vector $\eta$ evolves according to Eq.~\ref{eq:etaevol}.
The posterior and prior estimates $\hat{y}$ and $\hat{y}^-$ are computed using
Eqs.~\ref{eq:yhat}.

(5) For later learning of $\tilde{F}$, once $Z$ has been learned well enough to 
yield estimates $\hat{y}$ that are 
comparable or superior to the raw measurements $y$:
Eq.~\ref{eq:Ftil1} yields
\begin{equation}
\tilde{F}_t = \tilde{F}_{t-1} 
- \gamma_F \langle \eta_t \hat{y}'_{t-1} \rangle_p ~~.
\label{eq:Ftil1learn}
\end{equation}

The sets of computations (1) and (2) above can be performed in either order.
We will refer to the network's mode of operation during computation (1) as the
`initial $\tilde{F}$ learning' mode, and to that during (2) as the `offline sensor'
mode for learning $R$.
Once $R$ and 
$\tilde{F}$ have been approximately learned, the network's mode of operation
switches to a third, `Kalman,' mode, in which the network performs  
all of the sets of computations (3), (4), and optionally (5), at each time step $t$.  
The transitions between the various operating modes are effected by 
controlling the 
signal flows as described below in connection with Fig.~\ref{1a}.

\subsection{Neural circuit and flow diagram}

We now describe the neural circuit and set of signal flows that follow 
naturally from the above set of equations, and that implement Kalman estimation
and system identification. 

There are two connection 
matrices, $R$ and $Z$ (or $Z^{-1}$), whose learning rules
have a Hebbian term of the form $vv'$
(rather than $vz'$ with $v \neq z$); we therefore implement these
as two sets of lateral connections, each within its own layer.

\begin{figure}[htb]
\includegraphics[width=6in]{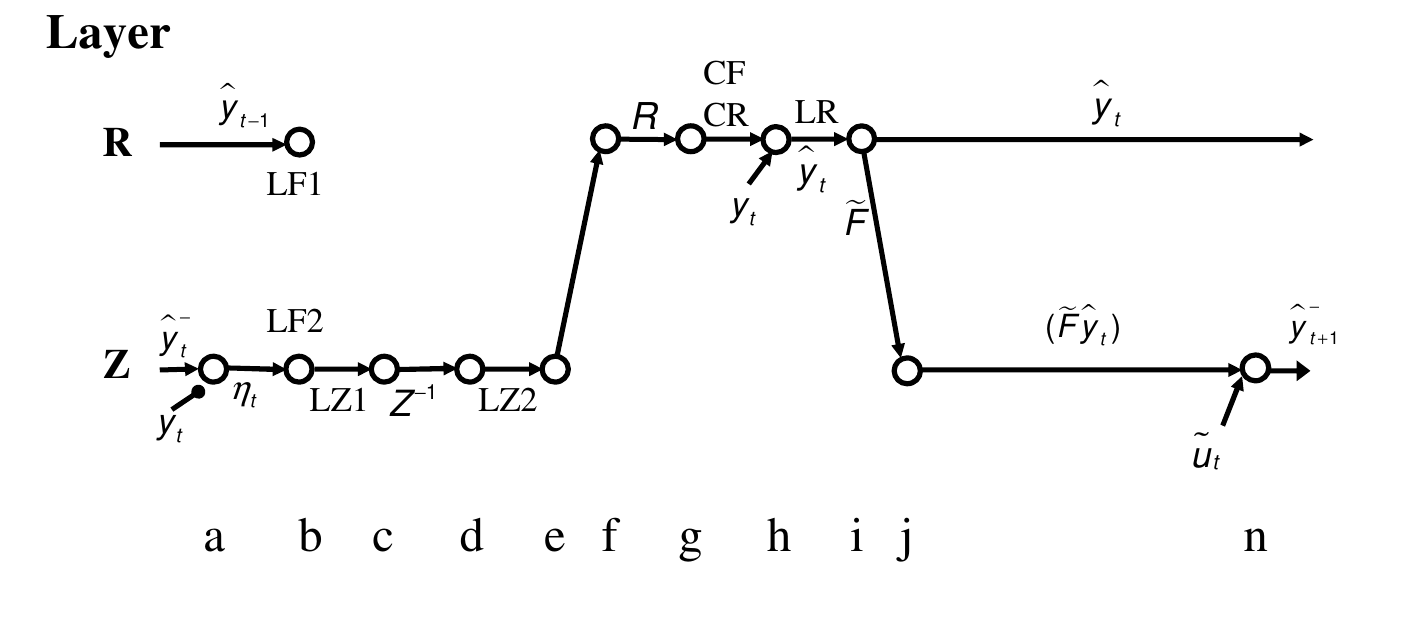}
\caption{ 
Layered organization of signal flow (a two-layer recurrent NN)
implied by the
NN algorithm for
Kalman estimation (filtering and prediction) and system identification.
All unfilled circles within a single layer denote the same set of $D_{\rm y}$
nodes in that layer, at each of many microsteps (time `ticks') denoted 
a, b, $\ldots$, n.
Computation proceeds from left to right along the links, for a single value of 
the time index $t$.
At extreme right, $t$ is incremented, 
and signal flow resumes at extreme left. 
Where two inputs enter an unfilled circle, the input activity vectors are 
combined;
arrowhead inputs are added and filled-circuit inputs are subtracted. 
Links labeled by a matrix denote that the activity vector at that link is 
multiplied by that matrix of connection strengths.
Link labels starting with C and L denote 
where signal flow is cut (`C'), and where signals are used for matrix learning (`L'),
during specific modes of operation (see text).
}
\label{1a}
\end{figure}

We therefore 
consider a recurrent NN having, for now, two layers (denoted R and Z)
with $D_{\rm y}$ nodes in each 
layer.\footnote{
If $N_{\rm feat} > 1$, one type of hardware 
implementation would have, within each layer,
$N_{\rm feat}$ sets 
of $D_{\rm y}$ nodes each (one set for each $p=1, \ldots, N_{\rm feat}$),
each set functioning independently of the others
except for an averaging process (over $p$) as in Eq.~\ref{eq:Mlearn}.
} 
(Note that $D_{\rm y}$, the 
dimensionality of the vector $y$, is also
the dimensionality of $\hat{y}$, $\hat{y}^-$, $\eta$, and $\tilde{u}$). 
Each node computes a linear combination of its real-valued scalar inputs.
There are three weight matrices to be learned: $\tilde{Z} (\equiv I-cZ)$ or $Z^{-1}$, 
$R$, and $\tilde{F}$.
The network's function is controlled so that computational steps occur,
and inputs are presented to nodes,
in a prescribed sequence.
(E.g., in a hardware implementation, a connection pathway might be enabled or
disabled, affecting signal flow and processing.)
On a `macro' time scale, each major time step $t$ corresponds to 
the presentation of 
a new measurement vector $y_t$ (and optionally $\tilde{u}_t$) and the computation 
of the one-step-ahead 
prediction $\hat{y}^-_{t+1}$.
On a faster, `micro,' time scale, we break each major time step into
multiple substeps (each called a `tick') denoted by lowercase letters
in alphabetical order.  See Fig.~\ref{1a}.

In this paper we assume that, for a hardware NN implementation,
the necessary apparatus is provided to control the signal 
flows in the various modes, but we do not discuss such 
apparatus explicitly.  The same is true of the apparatus
for sequencing the `ticks.' 

\subsubsection{`Kalman mode' of operation}

We first trace the signal flows for the computation of 
the execution Eq.~\ref{eq:etaevol} using   
Fig.~\ref{1a}, which depicts the ticks within time step $t$ from left to right. 
The computation of Eq.~\ref{eq:etaevol} 
is part of `Kalman mode' operation, defined above.
We will later discuss the learning and initialization during this mode.
Each (unfilled) circle denotes the set of $D_{\rm y}$ nodes in its layer, 
and each operation 
on this set of nodes is a matrix addition or subtraction  
of two input vectors to that circle, 
or a matrix multiplication 
of an input vector by
the indicated weight matrix.
At the left edge (during tick a), the new measurement input $y_t$ and the 
prediction
$\hat{y}^-_t$ (which was made at the end of the previous time step)
are combined in layer Z to form $\eta_t = \hat{y}^-_t - y_t$ 
(Eq.~\ref{eq:eta}).
The activity vector of the layer-Z nodes remains equal to 
$\eta_t$ until tick c.
Then the lateral connections between pairs of nodes in layer Z 
(which include self-connections) are enabled, and
the resulting activity vector at tick d is $Z^{-1} \eta_t$
(as discussed in the next paragraph). 
This activity vector is transported to layer R (between ticks e and f).
The lateral connections in layer R
multiply this activity vector by the matrix $R$, yielding 
$R Z^{-1} \eta_t$ at tick g.
The measurement vector $y_t$ is added at tick h to
yield $\hat{y}_t = y_t + R Z^{-1} \eta_t$ (Eq.~\ref{eq:yhat}).
This vector is 
multiplied, between ticks i and j, 
by the connection strength matrix $\tilde{F}$ of 
feedback connections from layer R to layer Z.
A control input vector $\tilde{u}_t$ is optionally added at tick n
(the tick letters skipped here are reserved for later use),
to yield $\hat{y}^-_{t+1}$ (Eq.~\ref{eq:yhat}), which is
the prediction of $Y_{t+1}$, the
`ideal noiseless measurement' at time $t+1$.
The right edge of Fig.~\ref{1a} is understood to loop back to the
left edge as the time index is advanced by one step;
thus the activity vectors $\hat{y}_t$ in layer R and 
$\hat{y}^-_{t+1}$ in layer Z
at tick n are relabeled $\hat{y}_{t-1}$ and $\hat{y}^-_t$,
respectively,
at tick a of the next time step.  

How are $\tilde{Z}$ or $Z^{-1}$ learned, and $\tilde{F}$ refined, during 
Kalman mode?  
If the lateral connections in layer Z are to embody the weight matrix 
$\tilde{Z}$
(Method 1 of section 3.3)
then Hebbian learning at link LZ1 (between ticks b and c) implements
Eq.~\ref{eq:Dlearn} using the activity vector $\eta_t$.  The iterative 
computation of Method 1 then produces 
the output $Z^{-1} \eta$ for each $\eta$, using the connections $\tilde{Z}$.
Alternatively, if the lateral connections are to embody $Z^{-1}$
(Method 2), 
then Hebbian learning at link LZ2 implements Eq.~\ref{eq:Zinvlearn},
using $(Z^{-1})_t \eta_t$, which is 
the activity vector at tick d
after one pass of $\eta_t$ through the 
$Z^{-1}$ connections.   
Finally, $\tilde{F}$ updating is performed by Hebbian learning 
across the $\tilde{F}$ connections 
between layers R and Z at tick b
(indicated by the labels LF1 and LF2 in Fig.~\ref{1a}).
The activities at the two sets of nodes 
at tick b
are $\hat{y}_{t-1}$ and $\eta_t$,
yielding Eq.~\ref{eq:Ftil1learn}.  (That is, 
the $\tilde{F}$ connections are used at tick b for updating $\tilde{F}$.
Since these connections are 
being used to multiply an activity vector by $\tilde{F}$ 
only between ticks i and j, but not at tick b,
no line is drawn between the layers at tick b in Fig.~\ref{1a}.)  

At the beginning of `Kalman mode,' $\tilde{Z} \equiv I-cZ$ is initialized 
by taking $Z$ to be an arbitrary
positive-definite symmetric matrix.  If $N_{\rm feat}$ is sufficiently large,
it can be convenient (and in keeping with the initialization given below
Eq.~\ref{eq:etaevol} and used in Appendix A) to choose
$Z_0 = \langle \eta_0 \eta'_0 \rangle_p$.  
This choice is not required in practice, however. 

\subsubsection{`Offline sensor' mode for learning $R$}

In this mode we cut off signal flow at the line labeled CR 
(`C' denotes `cutoff') in layer R 
between
ticks g and h, and we also cut off input from the external plant, 
so that the sensors
are running `offline' and they provide only measurement (sensor) noise
to the network at the input labeled $y_t$ at tick h; 
i.e.,
$y_t = n_t$ in Eq.~\ref{eq:yofx}.  
Then the activity vector in layer R at the line labeled LR 
(`L' denotes `learning') 
between ticks h and i
is just $n_t$, and Hebbian learning (at link LR)
of the lateral connection strengths within layer R 
yields $R \approx E(n_t n'_t)$.  
No further processing is done during 
this mode of operation.  
(When $\gamma_R < 1$ in Eq.~\ref{eq:Rlearn}, it is convenient to 
initialize $R$ to the zero matrix.) 

\subsubsection{`Initial $\tilde{F}$ learning' mode}

In this mode, we cut off signal flow at the line in layer R
labeled CF, between ticks g and h.
Then the activity vector in layer R is $y_t$ (rather than $\hat{y}_t$)
after tick h of the current time step, and remains so until 
tick b of the next time step, where it is called 
$y_{t-1}$ since $t$ has been incremented by one.
In layer Z, the activity after tick n is $\tilde{u}_t + \tilde{F} y_t$,
so the activity at tick b of the next time step is
$\tilde{u}_{t-1} + \tilde{F}_{t-1} y_{t-1} - y_t = \epsilon_t$ 
(defined just before 
Eq.~\ref{eq:Ftil0}).
Thus the Hebbian learning rule at step b uses the 
activity vectors that are on line LF1 (layer R) and line LF2 (layer Z) to
update the connection matrix $\tilde{F}$ between those layers, and the rule  
is as given by Eq.~\ref{eq:Ftil0learn}.  

$\tilde{F}$ may be initialized to be an arbitrary $D_{\rm y} \times D_{\rm y}$ 
matrix.

\begin{figure}[htb]
\includegraphics[width=3in]{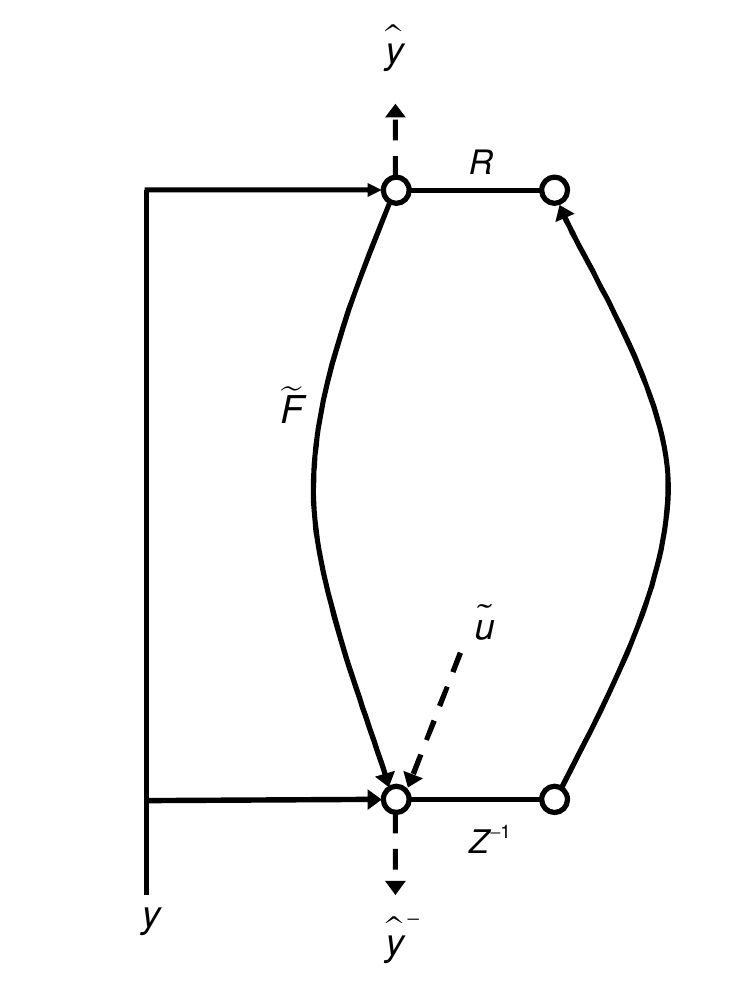}
\caption{ 
The NN circuit required to enable the signal flow of
Fig.~\ref{1a}.
Within each layer, the pair of circles denotes the set of $D_{\rm y}$ nodes,
and the labeled line joining them denotes the weight matrix 
(or its inverse)
of the lateral connections in that layer.
Dashed lines denote optional input or output connections.
Additive and subtractive inputs are not distinguished here; 
both are indicated by arrowheads.  
}
\label{1b}
\end{figure}


\subsubsection{Circuit diagram}

The static neural circuit for Kalman estimation and system identification --
showing all connections, but omitting the explicit time flows -- is shown
in Fig.~\ref{1b}.  The signal flows detailed in Fig.~\ref{1a} can easily 
be traced through this circuit.  Optional input $\tilde{u}$ and optional
outputs $\hat{y}$ and $\hat{y}^-$ are indicated by dashed lines.
(Figures \ref{1a} and \ref{1b} are, apart from minor modifications, subsets of 
Figs.~\ref{3ab} and \ref{3c}, which depict the neural circuit diagram for the fully integrated
algorithm, comprising Kalman control 
as well as estimation and system identification.
For a discussion of the 
relation between the signal flows and the static circuit for the 
full algorithm, as well as 
a summary block diagram of the full algorithm,
see section 5.1 
and Appendix B.)

\subsection{Numerical example}

\begin{figure}[htb]
\includegraphics[width=6in]{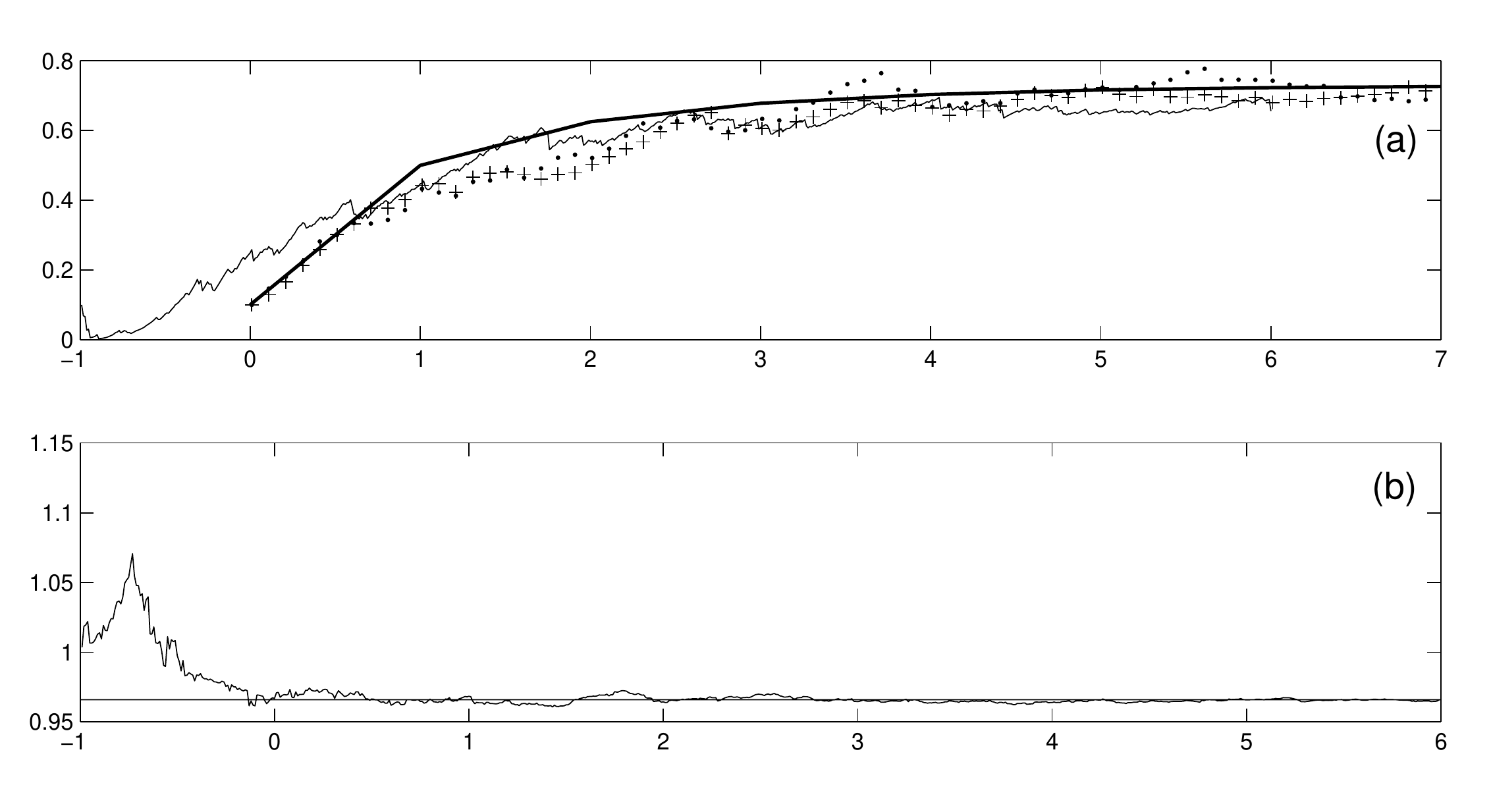}
\caption{ 
(a) Example of classical and neural Kalman filter learning.  
The (2,2) component of the 
$2 \times 2$ matrix $(I-HK_t) = RZ_t^{-1}$
(see text just before Eq.~\ref{eq:Zevol})
is plotted vs.~time step $t$,
for each of four computational methods.  See
text (section 3.6)
for details.
(b) Learning of (2,2) component of $\tilde{F}$ vs.~$t$, starting 
with arbitrary matrix.  
Horizontal line denotes true value of $\tilde{F}_{22}$.
}
\label{2}
\end{figure}

Numerical simulations (Fig.~\ref{2}) illustrate that the results of the NN estimation 
(KF) algorithm 
agree with the classical KF matrix 
solution, apart from 
fluctuations that decrease (not shown) as one increases the sample size used to  
estimate the covariance of $\eta$ at each time step. 
By contrast, recall that the classical Kalman algorithm 
is given the exact plant state covariance $Q$ and uses 
matrix-times-matrix operations, 
not available to the NN, 
to compute the covariance of the estimation error.

For our example, we consider a two-dimensional plant state.
The plant and measurement processes are defined by 
the following parameters 
(see 
section 2
for definitions): $F$ and $H$ are 2-d 
counterclockwise
rotations about the coordinate origin
by $15^o$ and $50^o$ respectively, and the plant and noise covariance matrices are
$Q = 10^{-5} I$ and $R = 10^{-4} I$ respectively, where 
$I$ is the $2 \times 2$ identity matrix.  We take $\tilde{u}_t \equiv 0$.

The learning rates are preferably allowed to be 
time-varying for more efficient learning.
Here, the learning rates 
$\tilde{\gamma}_Z \equiv \gamma_Z / N_{\rm feat}$ and (in run 4 below)
$\tilde{\gamma}_F \equiv \gamma_F / N_{\rm feat}$ are
adaptively adjusted using the method of 
Murata et al. (1997). 
In their notation, the values of the rate control parameters, 
which we have made no attempt to optimize, are 
$\{ \alpha, \beta, \gamma, \delta \} = \{ 0.5, 30, 0.05, 0.1 \}$
for $Z$, and $\{ 0.1, 3, 0.05, 0.04 \}$ for $\tilde{F}$.  

In Figure \ref{2}a, 
the (2,2) 
component of the $2 \times 2$ matrix $(I-HK_t) = R Z_t^{-1}$ 
is plotted vs.~time step $t$, for four different runs.  
All runs start with the same arbitrary $(I-HK_0)$.  The learning of the 
measurement noise covariance matrix $R$ is not shown here; 
$R$ is assumed already known.

Run 1 (thick solid curve): The classical Kalman 
Eqs.~\ref{eq:KPm} are run for 7 time steps, and the resulting values
are connected by straight line segments.  
Note that the results are identical when the transformed matrix 
Eq.~\ref{eq:Zevol} is used in place of Eqs.~\ref{eq:KPm}.

Run 2 (dotted curve): The neural network KF algorithm derived in this paper is used. 
In this run, one feature (a measurement vector $y$) is tracked for 700 time steps; 
i.e., $N_{\rm feat} = 1$.  
Here the displayed time scale is compressed 100-fold.
Every tenth value is plotted for better readability.

Run 3 (curve denoted by `+' signs): The same NN algorithm, but tracking 
$N_{\rm feat} = 100$ 
independent and simultaneously tracked
features for 7 time steps.  
[Here, we update $Z$ incrementally for  
each feature $p = 1, \ldots, N_{\rm feat}$,
rather than batch-averaging all $p$, at a given time step. 
Thus $N_{\rm feat}$ values of $(I-HK)$ are computed during each unit 
time interval.]  Every tenth value is plotted.  

Runs 1-3 are all computed using the true 
fixed value of $\tilde{F}$.  
 
Run 4 (thin solid curve): Same as for run 3, but here $\tilde{F}$ is 
initially arbitrary and
is learned from the measurement stream.  In the plot of this run, the time values 
are left-shifted by one unit relative to the curves for runs 1-3, 
in order to compensate for the startup time required to learn $\tilde{F}$
approximately.  
Referring to the values $t_{\rm plot} \equiv t-1$ on the abscissa of
this time-shifted plot:
We update $\tilde{F}$ 
using the raw measurement values $y$ from $t_{\rm plot}=-1$ to $0$,
because $Z$ is initially arbitrary and cannot yet give useful estimates.
For $t_{\rm plot}>0$, we update $\tilde{F}$ using the estimates 
$\hat{y}$ that depend on $Z$.  We start updating $Z$ at $t_{\rm plot}=-1$, 
even though this update uses values of $\tilde{F}$ that are initially
arbitrary and not yet reliably learned.  (For additional technical detail
regarding runs 3 and 4, see the last paragraph of this section.) 

The learned value of the (2,2) component of $\tilde{F}$, from run 4, is 
plotted vs.~$t_{\rm plot}$ in
Fig.~\ref{2}b.
Note that
the observations used to learn $\tilde{F}$ 
must span a sufficient portion of the 
dynamical space for learning to be adequate.
For example, if the measurement vectors $y$ used to 
learn $\tilde{F}$ were to have values 
that span a significant range only in their first component, 
then the (1,1) component of $\tilde{F}$ would be well learned, but
the (2,2) component would not.     
By way of contrast, as noted in section 2, classical 
Kalman estimation assumes that the true $F$ 
and $H$ have been specified to the algorithm. 

Additional technical details regarding runs 3 and 4:
Run 4 is generated by defining, at the starting time $t=0$ and 
for $1 \leq p \leq N_{\rm feat}$:
$\tilde{F}_0(p) \equiv \tilde{F}_0$, 
$Z_0(p) \equiv Z_0$, 
and $\hat{y}_0(p) \equiv y_0(p)$,
where $\tilde{F}_0$ and $Z_0$ are arbitrary 
matrices and $y_0(p)$ is the measurement vector of the
$p$th feature at $t=0$.
The matrices are then incrementally learned as follows:

For $t=1, 2, \ldots$: 

~~~~~For $p=1, 2, \ldots, N_{\rm feat}$, calculate:

\begin{eqnarray}
\eta_t(p) & = &
\tilde{F}_t(p-1) \hat{y}_{t-1}(p) - y_t(p)~~; \nonumber \\
\tilde{F}_t(p) & = & \tilde{F}_t(p-1) 
- \tilde{\gamma}_F \eta_t(p) y'_{t-1}(p)~~; \nonumber \\
Z_t(p) & = & (1-\tilde{\gamma}_Z) Z_t(p-1) + 
\tilde{\gamma}_Z \eta_t(p) \eta_t(p)~~. 
\label{eq:numer}
\end{eqnarray}
(Run 3 is generated in the same way as run 4, except without
$\tilde{F}$ learning.)
Here $\tilde{F}_t(p-1)$ is understood to mean 
$\tilde{F}_{t-1}(N_{\rm feat})$ when $p=1$, and similarly for
$Z$, $\hat{y}$, and $y$.  
These three equations correspond, respectively, to:
Eqs.~\ref{eq:yhat} and \ref{eq:eta}; Eq.~\ref{eq:Ftil0learn}
when $t=1$ or Eq.~\ref{eq:Ftil1learn} when $t>1$;
and Eq.~\ref{eq:Zlearn}.
That is, at each $t$, the algorithm cycles through 
the features one at a time, updates $\tilde{F}$ and $Z$,
then uses the most recent values of $\tilde{F}$ and $Z$ 
to perform the update for the next feature.
The first of Eqs.~\ref{eq:numer} differs subtly from 
Eqs.~\ref{eq:yhat} and \ref{eq:eta}, since it uses
$\tilde{F}_t(p-1) \hat{y}_{t-1}(p)$ for more efficient computation,
rather than 
$\tilde{F}_{t-1} \hat{y}_{t-1}(p)$.

\section{Neural algorithm for optimal Kalman control}

As we did above for Kalman estimation, we first transform the classical Kalman
control
equations, then show how to implement them in an NN.  The NN algorithm
and signal flows for Kalman control are integrated into those derived above for
Kalman estimation and system identification.    

To pass from Eqs.~\ref{eq:LS} to new equations in measurement space,
we define the transformations: 
\begin{eqnarray}
\tilde{g} \equiv H'^+ B'^+ g B^+ H^+ ~~; & &
~~ \tilde{r} \equiv H'^+ r H^+ ~~; \nonumber \\
\tilde{L}_{\tau} \equiv -HB L_{\tau} H^+ ~~; & &
~~ T_{\tau} \equiv H'^+ S_{\tau} H^+  +  \tilde{g} ~~.
\label{eq:KCtransf}
\end{eqnarray}

The transformed matrix equation that corresponds exactly to Eqs.~\ref{eq:LS}
is then
\begin{equation} 
T_{\tau-1} = 
\tilde{F}' \tilde{g} (I - T_{\tau}^{-1} \tilde{g} ) \tilde{F} + 
\tilde{r} + \tilde{g} 
\label{eq:Tevol}
\end{equation}  
(see Appendix A.2 for proof).
Similarly to the case of 
NN Kalman estimation, we will represent $T_{\tau}$ as the covariance
of the distribution of a stochastic
activity vector $w_{\tau}$, so that the learning rule for 
$T_{\tau}$ may
be recast as an evolution equation for $w_{\tau}$.  However, whereas
the physically meaningful quantity $\eta_t$ 
was the vector whose covariance 
equaled $Z_t$ in the case of estimation, we now have to construct $w_{\tau}$
from terms that are based on the goal of the control problem, i.e., the cost
function to be minimized. 

We introduce the activity vector $w_{\tau}$, and construct a rule for 
computing $w_{\tau -1}$ in terms of $w_{\tau}$,
such that $E(w_{\tau}w'_{\tau})$ satisfies the same evolution equation as 
$T_{\tau}$ (Eq.~\ref{eq:Tevol}):    
\begin{equation} 
w_{\tau -1} = 
- \nu^g_{\tau -1} + 
 \nu^r_{\tau} +  \tilde{F}'
 ( \nu^g_{\tau} + \tilde{g} T_{\tau}^{-1} w_{\tau} ) 
\label{eq:wevol}
\end{equation} 
(see Appendix A.2 for proof).
Here $\nu^g_{\tau}$ and $\nu^r_{\tau}$ are random vectors, or 
internally generated `noise,'
drawn from distributions having mean zero and covariances $\tilde{g}$ and 
$\tilde{r}$ respectively. 
These noise generators are the means by which the neural network
represents the cost matrices $\tilde{g}$ and $\tilde{r}$. 
Note that we have a learning rule for $T_{\tau}$, but need to 
compute $(T^{-1})_{\tau} w_{\tau}$ in Eq.~\ref{eq:wevol}; thus $T$ here plays
the role analogous to $Z$ in neural Kalman estimation, and the present computation 
is handled using the same methods (see next paragraph).   

Learning: At the current time $t_0$, 
a set of KC matrices $T_{\tau}$ to be used at future times is learned
by iteratively computing Eq.~\ref{eq:wevol}  
for $\tau = N, N-1, \ldots, t_0 +1$,
starting
with $w_N = \nu^r_{N+1} - \nu^g_N$
(corresponding to $S_N = r$). 
In the idealized limit in which the sample average converges to the 
expectation value for the $w_{\tau-1}$ distribution -- i.e.,  
$\langle w_{\tau-1} w'_{\tau-1} \rangle \rightarrow E(w_{\tau-1} w'_{\tau-1})$ --
neural control learning using 
$T_{\tau-1} = \langle w_{\tau-1} w'_{\tau-1} \rangle$ 
exactly yields optimal Kalman control. 
In practice, we use either Method 1 of section 3.3 
above,
using the update rule for $(I-cT)$ (cf.~Eq.~\ref{eq:Zlearn}) where 
$T_{\tau-1} = (1-\gamma_T) T_{\tau} 
+ \gamma_T \langle w_{\tau} w'_{\tau} \rangle_p$; 
or Method 2 (cf.~Eq.~\ref{eq:Zinvlearn}), using 
$(T^{-1})_{\tau-1} = (1+\gamma_T) (T^{-1})_{\tau} 
- \gamma_T \langle w_{\tau} w'_{\tau} \rangle_p$
with the same caveats
that were described for 
$Z^{-1}$ in section 3.3.  
Either method yields an approximation to the optimal 
Kalman controller that is limited in accuracy only by the 
deviations between sample averages and expectation values.
(Unlike the case of Kalman estimation, in which $N_{\rm feat}$ is limited by the
number of available features to be tracked, here the $w_{\tau}$ values are 
internally computed, so one can in principle use arbitrarily many independent
$w_{\tau}(p)$ values
at each value of the time index $\tau$.)

Execution: For any or all of the time  
steps $t = t_0, \ldots, N$,
the $T_t$ or $(T^{-1})_t$ computed above can now be used 
to compute the desired control signal
\begin{equation} 
\tilde{u}_t = \tilde{L}_t \hat{y}_t 
= (-I + T_t^{-1} \tilde{g} ) \tilde{F} \hat{y}_t ~.
\label{eq:u}
\end{equation}

For a software implementation, it is most convenient to store 
the sequence of $T$ or $T^{-1}$ matrices during learning, 
and retrieve them 
in reverse order during execution.
In hardware, one can choose either to 
(a) store and retrieve as above    
(possibly using different parts of a larger NN to store each matrix, 
not discussed here), 
or 
(b) retain only the last-computed matrix $T_{t_0}$, use it for execution 
at the current time $t_0$, then
relearn the $T$ matrices at the next time step $t_0+1$.
Alternatively, one may (c) approximate ideal KC by  
using the same computed matrix for several time steps.

\begin{figure}[htb]
\includegraphics[width=6in]{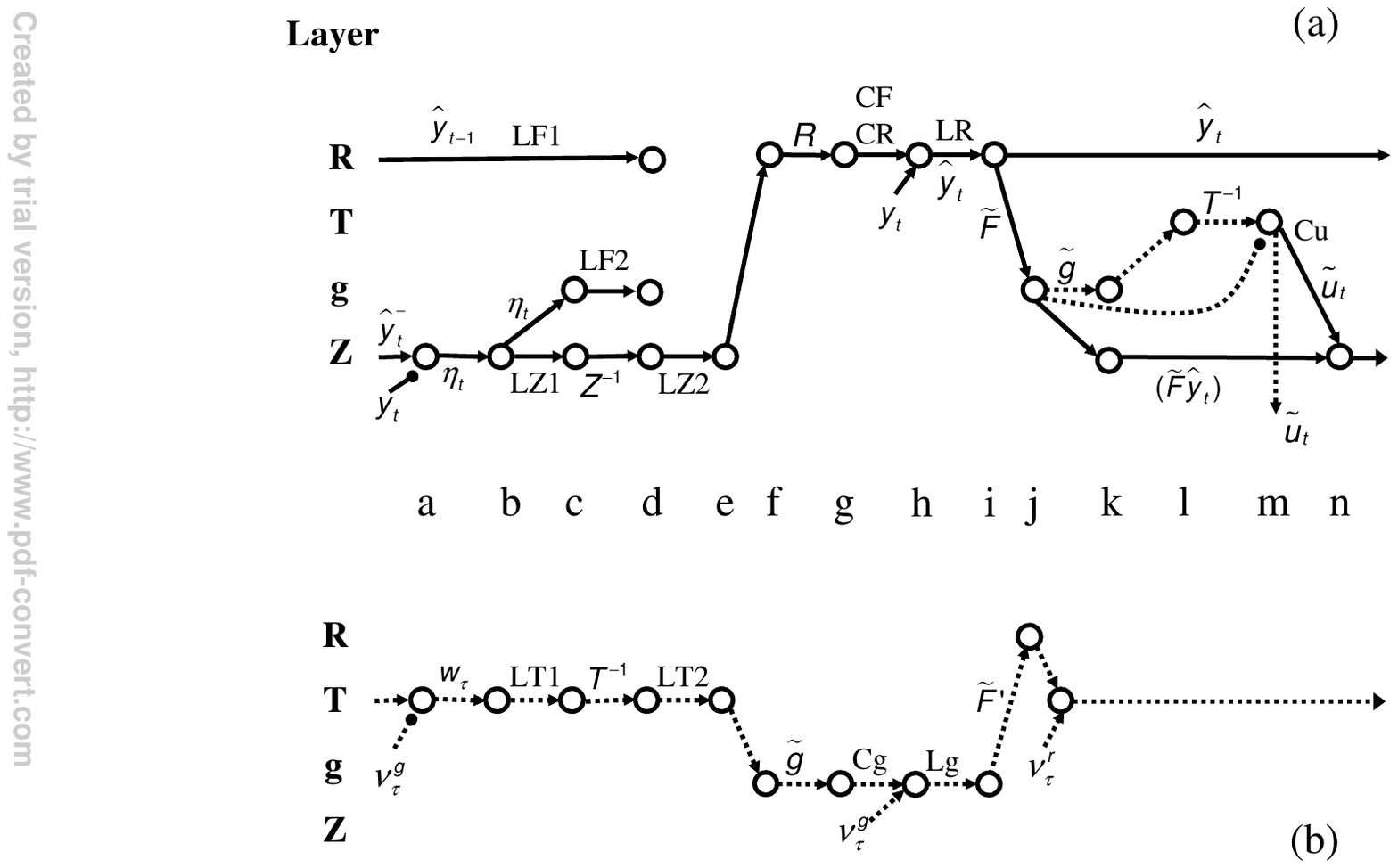}
\caption{ 
Layered organization of signal flow required by full NN algorithm
for Kalman estimation, system identification, and Kalman control. 
(a) Signal flow for system identification and for 
learning and execution of Kalman estimation (solid links), 
and for execution of Kalman control (dashed links), shown in a four-layer 
recurrent NN, all at time $t$.
Other notation as in Fig.~\ref{1a}. 
(b) Signal flow for learning of Kalman control (dashed links), 
shown separately in the same 
four-layer NN, all at time index $\tau$ of the KC learning process.   
At extreme right, $\tau$ is
decremented, and signal flow resumes at extreme left. 
Other notation as in Fig.~\ref{1a}.
}
\label{3ab}
\end{figure}

To show how the  
learning and execution of Kalman control (KC) 
is added to the neural network, we
refer to Figs.~\ref{3ab} and \ref{3c}.  Figure \ref{3ab}a (for KC {\it execution}) 
is essentially Fig.~\ref{1a} augmented by two additional layers, denoted
T and g (the latter label 
not to be confused with `tick g' on the horizontal axis),
and with additional processing 
(indicated by the dotted lines between ticks j and n) to compute $\tilde{u}_t$
using the new weight matrices $T$ (or $T^{-1}$) and $\tilde{g}$.
The {\it learning} of these matrices is performed by the processing described 	 
in Fig.~\ref{3ab}b, which takes place
during different modes of operation (to be described) than does the 
processing described by Fig.~\ref{3ab}a.  
Since KC learning involves some special features not encountered in KF learning,
we first consider Fig.~\ref{3ab}a, which embodies KF learning and execution, combined with 
KC execution (but not KC learning).

The new processing begins 
in layer g at tick j.
Note that the $\tilde{F}$ connections now go from layer R to g (rather than to Z
as in Fig.~\ref{1a}).
The activity vector at layer g and tick j is thus $\tilde{F} \hat{y}_t$, and 
this activity is passed on unchanged to layer Z at tick k.
From tick j to tick m, the activity vector computed by the two dotted-line signal 
flows is
$\tilde{u}_t = ( T^{-1} \tilde{g} - I ) \tilde{F} \hat{y}_t$
(as required by Eq.~\ref{eq:u}).
Since $\tilde{g}$ and $T$ (or $T^{-1}$) are the weights on two 
different sets of lateral connections, we have assigned 
each of these connection matrices
to a different layer (g and T respectively).    
The vector $\tilde{u}_t$ is provided as output from the network to effector `organs'
(which act on the external plant), and is also provided as
an `efferent copy' to the network itself (at tick n), where it is used to compute
the prediction $\hat{y}^-_{t+1}$.

Unlike the case of KF, where $Z$ or $Z^{-1}$
is learned as part of the same process that 
computes the predictions $\hat{y}^-$, here the KC {\it learning} process 
(updating of $T$ or $T^{-1}$, and of $g$) 
must be done separately from KC {\it execution} (computation of $\tilde{u}$).
The same four layers are used for both KC learning and execution, but at different
times and in different modes.  
Thus, at a particular value of the execution time index $t=t_0$,  
the network processing mode is changed from 
`Kalman mode' (which now includes KC execution)
to `KC learning mode,' in which  
a sequence of steps for the KC learning 
time index $\tau = N, N-1, \ldots, t_0+1$ is
performed.  See Appendix B for further discussion. 

The KC learning process is described in Fig.~\ref{3ab}b.  The time step is 
labeled $\tau$,
and this label is {\it decremented} by one when we pass from tick n to tick a. 
The shift in mode from Kalman mode to KC learning mode 
can either be made at each 
time step $t$ [as in implementation (b) following Eq.~\ref{eq:u}],
or at only some time steps $t$ if means are provided to store the
computed values of $T$ or $T^{-1}$ [as in implementations (a) and (c) above]. 

Learning of $\tilde{g}$ is analogous to that of $R$:  The signal flow is 
cut off at link Cg (before tick h), so that the only input to the layer-g nodes
at tick h is the structured noise term $\nu^g_{\tau}$.  Then Hebbian learning
at the lateral layer-g connections implements 
$\tilde{g} \approx E[ \nu^g_{\tau} (\nu^g_{\tau})' ]$.  
No further processing occurs during this mode.     

For $T$ (or $T^{-1}$) learning, the entire signal flow pathway of Fig.~\ref{3ab}b is active.
Starting at tick a, 
the activity vector $w_{\tau}$ yields the following 
activity vectors at subsequent ticks:
\begin{enumerate}
\item $T^{-1} w_{\tau}$ at tick d;
\item $\tilde{g}T^{-1} w_{\tau}$ at tick g;
\item $\nu^g_t + \tilde{g}T^{-1} w_{\tau}$ at tick h;
\item $\nu^r_{\tau} + \tilde{F}' ( \nu^g_t + \tilde{g}T^{-1} w_{\tau} )$ at tick k;
\item $w_{\tau -1 } = -\nu^g_{\tau-1} + \nu^r_{\tau} + 
\tilde{F}' ( \nu^g_t + \tilde{g}T^{-1} w_{\tau} )$ at tick a of the new
time step $\tau-1$.
\end{enumerate}
The last equality comes from Eq.~\ref{eq:wevol}.

The activity vector $w_{\tau}$ is used to update $T$ or $T^{-1}$ 
in the same way that
$\eta_t$ was used to update $Z$ (Method 1) or $Z^{-1}$ 
(Method 2)
in the Kalman estimation algorithm (section 3.3).
Thus, using Method 1, Hebbian learning at link LT1 implements 
$T_{\tau} \approx E(w_{\tau} w'_{\tau} )$. 
Alternatively, using Method 2,
Hebbian learning at link LT2 implements
\begin{equation}
(T^{-1})_{\tau-1} = 
(1+\gamma_T) (T^{-1})_{\tau} - \gamma_T \langle [(T^{-1})_{\tau} w_{\tau} ] 
[(T^{-1})_{\tau} w_{\tau} ]' \rangle_p 
\label{eq:learnTinv}
\end{equation}
(subject to the same requirements on $\gamma_T$ and $T^{-1}$ symmetry
that were discussed in section 3.3 for the case of $Z^{-1}$ learning).
The learning is Hebbian,  
since the activity at tick d after one pass of $w_{\tau}$ through the 
$T^{-1}$ connections is $[(T^{-1})_{\tau} w_{\tau}]$.   

Note that the $\tilde{F}'$ connections in Fig.~\ref{3ab}b are 
shown as a signal flow 
line passing from layer g to R followed by a line 
from R to T, rather than passing directly from layer g to T.
This distinction
is irrelevant to a software implementation, but is shown here because 
$\tilde{F}'$ is learned using
a Hebbian rule that is identical to that used for $\tilde{F}$, and 
consequently 
it is convenient for $\tilde{F}$ and $\tilde{F}'$ to 
connect the same two layers (g and R), in opposite directions, in a hardware 
implementation.  $\tilde{F}$ and its transpose $\tilde{F}'$ may be thought
of as the same physical set of symmetric connections in an artificial hardware
implementation that allows bidirectional connections.  In a model 
that permits only unidirectional connections (in both directions), e.g., 
a model of
a biological network, $\tilde{F}$ and $\tilde{F}'$  
would be thought of as distinct sets of 
connections.  Even in the latter case -- 
since the same Hebbian rule is applied to both, with the same
pair of activities at the two ends of each pair of connections -- the two sets
will nonetheless 
learn matrices that are the transpose of each other, apart from differences
resulting from initialization, processing noise, and possibly 
faulty or missing connections.

\section{Discussion}

We have shown how to perform the learning and execution of 
Kalman estimation and control,
as well as system identification, 
using a neural network.  The method is asymptotically exact in the limit 
that 
certain sample averages of computed quantities
approach the corresponding expectation values over the distributions
of those quantities. 
The matrices $Z$ and $T$ that are iteratively computed, in order to learn and 
execute KF and KC respectively, are each equal to
the covariance of a distribution of computed activity vectors.  
In each case, 
vectors evolve over time 
via a sequence of transformations performed by the NN,
and are used, via Hebbian learning, to update a matrix of connection 
weights that represents 
the KF or the KC matrix.
The logical path of the derivation 
proceeds
from the classical Kalman solutions, 
to a transformed set of equations that involves only quantities measured by the  
NN, to a set of signal flows and computations, and finally 
to a layered NN architecture 
and circuitry
that supports those computations.  

Both the signal flows and the NN architecture 
appear to be
constrained by the requirements of the Kalman 
neural algorithm
(apart from small variations); that is,
they do not appear to represent
merely one among a large number of disparate possible choices of architecture
or signal flow.  
 
In the remainder of this section, we discuss
\begin{enumerate}
\item the considerations that constrain the architecture and signal flows;
\item applications to artificial NN design, and prior work;
and
\item resemblances between the derived architecture and biological 
networks in neocortex, and caveats 
involved in drawing inferences regarding biological
function.
\end{enumerate}  

\subsection{Constraints on NN architecture and signal flows implied by the
neural Kalman algorithm}

The assignment of activity vectors to distinct NN layers, and details of the 
signal flow among
layers, are significantly constrained by several requirements:
\begin{enumerate}
\item Each of the four matrices $R$, $\tilde{g}$, $Z$ (or $Z^{-1}$), and $T$ 
(or $T^{-1}$),
is learned using a Hebb rule that
contains a product of the form $vv'$; i.e., in each case the
activity vectors at the source and target ends of the connection matrix
are the same.  
(See Eq.~\ref{eq:Mlearn} with $z \equiv v$ for $R$ and 
$\tilde{g}$ learning; Eq.~\ref{eq:Dlearn} 
or \ref{eq:Zinvlearn} for $Z$ or $Z^{-1}$ learning; and the corresponding equations
for $T$ or $T^{-1}$ learning.)
When the Hebb rule is of this form, 
it is natural to implement the connection matrix as 
joining each node (of the set of $D_{\rm y}$ nodes) to each 
node (including itself) of the same set of nodes. 
Thus each such matrix describes a 
set of lateral connections, and is assigned to its own layer, denoted by 
R, g, Z, T in each of Figs.~\ref{3ab}a and \ref{3ab}b.  
\item For Hebbian learning, activity $\eta_t$ must be present at the input 
to the $Z$ (or $Z^{-1}$) connections; $w_{\tau}$ at the input to $T$ (or $T^{-1}$);
$y_t \equiv n_t$ at the input to $R$ during $R$ learning mode;
and $\nu^g_{\tau}$ at the input to $\tilde{g}$ during $\tilde{g}$ learning mode.
\item For Hebbian learning of $\tilde{F}$, activities      
$\hat{y}_{t-1}$ and
$\eta_{t}$ must be present simultaneously at the two ends of the $\tilde{F}$ matrix. 
Thus $\hat{y}_{t-1}$ must be held as the activity of one set of nodes in layer R
(of Fig.~\ref{1a} or \ref{3ab}a)
until $\eta_{t}$ has been computed at layer Z
and made available at layer Z (of Fig.~\ref{1a}) or layer g (of Fig.~\ref{3ab}a).
$\tilde{F}$ is {\it updated} at the 
time indicated by
links LF1 and LF2, but is {\it used} later in the signal flow, at the link 
labeled $\tilde{F}$.   
\item The transpose matrix $\tilde{F}'$ is required for KC learning 
(Fig.~\ref{3ab}b). 
$\tilde{F}'$ is learned by the same
algorithm, and at the same time, as $\tilde{F}$; thus it is assigned to connect 
the same two layers as $\tilde{F}$, but in the reverse direction.
Figure \ref{3ab} assigns $\tilde{F}$ to run from layer R to g, and 
$\tilde{F}'$ from g to R; 
this requires $\eta$ to be copied from layer Z to g as shown (just before LF2).
[As a slight variant, $\tilde{F}$ could run instead 
from R to Z; then $\eta$ would not need to be copied from Z to g,
but the solid path (Fig.~\ref{3ab}a) would require
a Z $\rightarrow$ g link just following the $\tilde{F}$ link from R to Z 
for KC execution, and the dashed path (Fig.~\ref{3ab}b) 
would require a g $\rightarrow$ Z link 
following link Lg,
to connect to $\tilde{F}'$.]   
\item The measurement vector $y_t$ is required twice as input: to layer Z, 
where it contributes 
to $\eta_t$, and to layer R, where it combines with $R Z^{-1} \eta_t$ to 
yield $\hat{y}_t$.
\end{enumerate}

Thus the arrangement shown in 
Fig.~\ref{3ab} is not an arbitrary way of laying out the NN algorithms we have derived; rather, the
four-layer organization and its signal flows appear to be
substantially determined (apart from small variations)
by the algorithms and the requirements of a NN implementation.  

\begin{figure}[htb]
\includegraphics[width=3in]{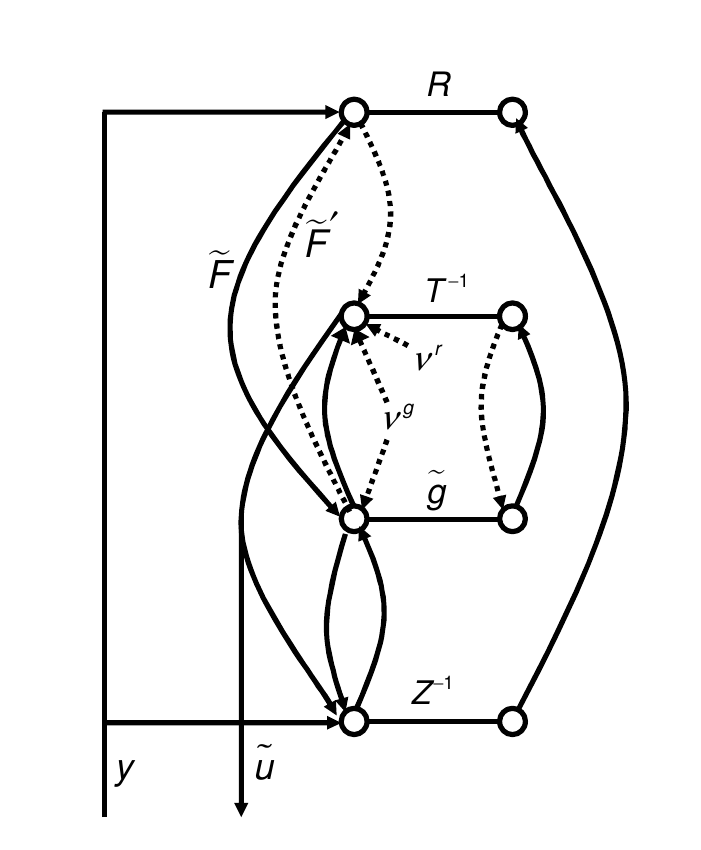}
\caption{ 
NN circuitry required to enable the signal flow of Fig.~\ref{3ab}.
Within each layer, the pair of circles denotes the set of $D_{\rm y}$ nodes,
and the labeled line joining them denotes the weight matrix 
(or its inverse)
of its lateral connections.
Dashed lines denote connections that are only involved in KC learning. 
Other notation as in Fig.~\ref{1b}.
}
\label{3c}
\end{figure}

The static neural circuit for the integration of Kalman estimation and control,
as well as system identification -- 
showing all connections, but omitting the explicit time flows
-- is shown in Fig.~\ref{3c}.  
The signal flows of Fig.~\ref{3ab} can be traced through this circuit 
(see Appendix B.1 as an aid).
The circuit operation comprises several modes (requiring 
appropriate functional switching) including: 
(a) learning of the $\tilde{F}$ and $\tilde{F}'$ connection weight matrices 
(system identification);
(b) normal `online' KF and execution of KC (`Kalman mode'); 
(c) iterative learning of the KC matrices $T$;
and
(d) initial or intermittent learning of matrices $R$ (for KF)
and $\tilde{g}$ (for KC).
Optional outputs $\hat{y}$, $\tilde{F}\hat{y}$, $\hat{y}^-$, and/or $\eta$ 
can be provided (not shown) from layers
R, g (or Z), Z, and Z (or g) respectively.

\subsection{Engineering applications and prior work}

As a mathematical 
and engineering 
method, these NN algorithms may prove useful for implementing 
estimation, control, and system identification in special-purpose hardware 
comprising simple computational elements,
especially with a large number of sets of such elements operating in parallel.
Even when the plant or measurement parameters change with time, the present 
NN algorithms 
learn the new dynamics automatically and converge to the new optimal solution 
after a transient period of adjustment. 
The well-known extended Kalman filter (EKF) (Haykin, 2001) 
and its variants yield 
approximate solutions for nonlinear plant and measurement processes, by repeatedly
linearizing the dynamics about an operating point.  Our NN algorithms likewise
yield approximate solutions in these cases, 
since the learned plant and measurement parameters are automatically
updated in response to the changing stream of noisy measurements (see text just below
Eq.~\ref{eq:Ftil1}).

In earlier work, 
neural networks have been used in conjunction with the Kalman filter (KF)
or Kalman control (KC) equations in several ways: 

The classical KF or EKF (extended Kalman filter) 
equations have been used to compute how the weights 
in a NN should be modified.  The NN weights 
to be determined
are treated as the unknown parameter values in a 
system identification problem, sets of input values 
and desired output values are specified, and the 
KF or EKF equations are used to determine the NN 
weights based on the sets of input and desired-output values.  
The equations are solved by means of conventional mathematical 
steps including matrix multiplication and inversion.  
That is, the weights are not computed or updated 
by means of a neural network.  Instead, the weights are 
read out from the neural network, provided as input 
arguments to the KF or EKF algorithm (which is not a NN
algorithm), the updated weight values are then provided as 
outputs of the KF or EKF algorithm, 
and the updated weight values are 
then entered into the neural network as the new 
weight values for the next step of the computation.  
For examples of this combined usage of a NN 
in conjunction with the classical 
KF or EKF equations, see 
(Haykin 2001, Rivals \& Personnaz 1998, Singhal \& Wu 1989, Williams 1992).
Rao \& Ballard (1997) also describe an EKF algorithm for weight update,
but do not address how the algorithm (with its matrix-times-matrix
computations and matrix inversion) could be implemented in a 
NN whose units have limited computational power.

The output from a nonlinear NN has been used in conjunction with that of a 
classical 
(non-neural) KF algorithm, to improve predictions when applied to a nonlinear 
plant process (Klimasauskas \& Guiver 2001, Tresp 2001).

A NN algorithm for KC (Szita \& L\H{o}rincz, 2004), 
using 
temporal-difference learning, performs KC
in the special case of stationary control, but 
is 
not applicable to the general KC case.
In the stationary control problem there is no specified 
time-to-target $N$; instead,
the time remaining to
the goal is either infinite, or is selected at each succeeding 
time step from a 
distribution that does not change with time. 

A recent KF-inspired NN algorithm (Szirtes, P\'{o}czos, \& L\H{o}rincz, 2005)
is described as a `neural Kalman filter.'
However, it substantially alters
Kalman's formulation, to the extent that
the resulting NN does not in general implement KF, even approximately. 
Although the initial prediction error (starting with an arbitrary prediction) 
is shown to decrease rapidly with time, 
this provides no 
evidence that the KF has been even approximately learned.  Indeed, 
a similar reduction in error is found even when an 
arbitrary, non-optimal, and unchanging filter,
differing greatly from the true KF, is used.  See Appendix C for 
details.

\subsection{Comparison with biology -- background}

Mammalian neocortex exhibits a significant degree of uniformity
in its layered architecture and pattern of interlaminar 
connectivity, although there are also well-known variations among
cortical areas (Mountcastle, 1998). 
A focus on the properties that are similar across cortical areas
and species
has led to models of neocortex that have, as a basic unit
on the 50- to 100-micron scale, the so-called local cortical circuit 
(LCC, minicolumn, canonical microcircuit)
(Callaway, 1998; Douglas \& Martin, 2004; Gilbert, 1983; Mountcastle, 1998).
The observed uniformity also
motivates a search for a set of core LCC processing functions that 
may be common to
sensory, motor, and other cortical 
areas, and that may enable the diverse functions of those 
areas (Grossberg \& Williamson, 2001; Poggio \& Bizzi, 2004). 

The blending of `bottom-up' sensory input and `top-down' 
model-driven expectations
has been discussed in the context of Bayesian inference and generative 
models, and various neural network (NN) algorithms  
are motivated by, approximate, or perform a portion of, the 
Bayesian inference 
process
(e.g., George \& Hawkins, 2005; Hinton \& Ghahramani, 1997;
Hinton, Osindero, \& Teh, 2006; Lee \& Mumford, 2003;
Lewicki \& Sejnowski, 1997; Rao, 1999; Rao, 2004; Rao, 2005; 
Todorov, 2005; Yu \& Dayan, 2005; Zemel et al., 2005).
The use of bottom-up and top-down signals in these algorithms has 
been noted to be reminiscent of the feedforward and 
feedback connections, both between different cortical areas and within the LCC. 
Bayes-optimal behavior has been found in 
human psychophysics 
experiments (e.g., K\"{o}rding \& Wolpert, 2004). 
Kalman filtering 
is well known to be, under certain conditions, 
an exactly solvable special (linear) 
case of Bayesian inference. 

I consider it plausible that the core functions of the LCC include
the prediction of future sensory input, the estimation of
noisy or missing input, and the generation of control
outputs, and that these functions are performed at 
multiple levels in sensory, motor, and other cortical areas.
Part of the motivation for the present work has been to explore
this conjecture.  To do this, it has appeared fruitful to ask: 
Is an implementation
of Kalman's methods for optimal estimation and control 
possible
within an artificial
neural network composed of 
simple processing nodes?
If so, what does such an implementation
appear to require of the
network's architecture and processing
-- how the network is divided into layers, the connections between 
and within layers, timing
considerations, etc.?
The results we have presented here suggest that requiring a
neural network to implement the Kalman
solutions imposes significant constraints on the network's form.
Because of this, resemblances that are observed between the artificial
and biological networks may have greater potential 
significance than they would have if the form of an artificial
Kalman neural network were quite unconstrained.  
However,  
any such comparison between the artificial and biological
networks involves many caveats, as we discuss next.

\subsection{Comparison with biology -- the neural Kalman circuit 
and the LCC}  

This subsection is necessarily speculative.
Here we show, first, that our KPC NN architecture, to which we 
have been led by the above constraints, bears certain resemblances to 
the observed architecture (layered organization, and the connectivity 
among layers, inputs, and outputs) of the LCC.  This is consistent with
the conjecture that the LCC's core functions include those of 
estimation (prediction and filling-in of missing or noisy data) 
and control, albeit in the context of nonlinear systems, interactions,
and feature discovery and analysis,
rather than in the simpler linear (or linearizable) Kalman context.

After identifying the resemblances, we turn to the differences between
the Kalman NN and the LCC, and to caveats that limit our current ability to
use the approximate Kalman-LCC `mapping' to
draw inferences regarding possible LCC function.

\subsubsection{Resemblances}

\begin{enumerate}
\item The KPC NN and the LCC are both recurrent 
neural networks.  Given the iterative nature of the 
classical Kalman algorithms, this is an unsurprising feature of 
the Kalman NN.  
\item The KPC NN has four layers (two if only 
Kalman estimation and system identification, but not control, are 
considered).  The LCC is typically
treated as having four layers (denoted as layers 6, 5, 4, and 2/3),
of which three (layers 6, 4, and 2/3) are considered important for 
sensory (as distinct from motor control) processing.
\item The `sensory' input to the KPC NN is required to enter at two
layers (denoted as layers Z and R).  Input to layer R 
(at tick h of Fig.~\ref{3ab}a) may be regarded
as primary, and that to layer Z as modulatory,  
in the sense that the raw measurement $y_t$ enters layer R
as the primary contribution to the computation of the estimate 
$\hat{y}_t$, while $y_t$ enters
layer Z (at tick a of Fig.~\ref{3ab}a) in order to compute the Kalman correction 
[which is $RZ^{-1}(\hat{y}^-_t - y_t)$] to the raw measurement.
Inputs to the LCC from `lower' levels of a sensory hierarchy (as
usually conceived)
are to layers 6 and 4, with the input to layer 4
considered as dominant, and that to layer 6 as modulatory.
In the LCC, unlike our linear NN, these inputs can interact nonlinearly.
\item The Kalman estimate of the present state, $\hat{y}_t$, is 
available as output from layer R, and the Kalman control signal $\tilde{u}_t$
from layer T.  
(The prediction of the state at the next time step, $\hat{y}^-_{t+1}$,
is also available from layer Z.)    
In the LCC, output to `higher' hierarchical levels 
is from layer 2/3, and that to `lower' levels is from layer 5.
For the LCC, the layer 2/3 output signals the
results of feature analyses 
(e.g., of features within a sensory `scene') 
that have been performed within that 
cortical area.  This is a nonlinear computation that can be 
considered analogous to linear Kalman estimation.
This putative LCC computation, and Kalman estimation, both yield
an improved knowledge of the external state, by suppressing noise
and `filling in' missing data, even though linear estimation is not
capable of inferring, or making `decisions' about, the presence or 
absence of particular features.  The layer-5 LCC output provides 
motor control signals -- again by a nonlinear process that 
has greater capability than, but can also
be considered analogous to, or a more powerful version of, Kalman control.   
\end{enumerate}  

We compare diagrammatically the interlaminar signal flow of the KPC NN with 
the putative principal signal flows (Gilbert, 1983)   
of the LCC.
The signal flow for Kalman estimation (learning and execution) 
and Kalman control (execution only) (Fig.~\ref{3ab}a) may be schematized, 
considering layers g and T as a unit,
as Dgm.~D1 (below, left): 
\begin{center}
\begin{tabular}{ccccccc|ccccccccc}
$\{ \;\;\; {\rm R} \;\;\; \}$ &  $\rightarrow$ & (g,T) & $\rightarrow$ & Z 
& $\rightarrow$ R & && 4	& $\rightarrow$ & 2/3 & $\rightarrow$ & 5	
&$\rightarrow$ & 6	& $\rightarrow$  4   \\
$\uparrow \;\;\;\; \downarrow$ & & $\downarrow$   & & $\uparrow$ & &&   
&  $\uparrow$ &  & $\downarrow$ & & $\downarrow$ & & $\uparrow$ &   \\
$y \;\;\;\; \hat{y}$ & & $\tilde{u}$ & & $y$ &  {\bf [D1]} & && in	
&  & ${\rm out}_S$ & & ${\rm out}_M$ & & in & 	\bf{[D2]} \\	
\end{tabular}
\end{center}
(The KC learning of Fig.~\ref{3ab}b adds a g $\rightarrow$ R path.) 
By comparison,
Gilbert's (1983) proposal 
for the principal LCC signal flow among 
the layers 
6, 5, 4, and 2/3 is shown in Dgm.~D2 (above, right).
More recent work is consistent with, and expands upon, this basic 
flow (Callaway, 1998; Douglas \& Martin, 2004; Raizada \& Grossberg, 2001). 
Layer 4 is elaborated in visual cortex 
and is much less prominent in motor than in sensory cortex,
while layer 5 is more prominent in motor cortex (Mountcastle, 1998) 
and provides motor control output 
(both in motor and in sensory cortex, 
e.g., from V1 to superior colliculus)
denoted here by ${\rm out}_M$.
Layer 2/3 integrates contextual inputs from outside the classical 
receptive field,
and provides output, denoted by ${\rm out}_S$, to other cortical 
areas that process
`higher-level' perceptual features.  

We consider, in the next subsection, 
the extent to which it is reasonable to take seriously 
these resemblances between the KPC NN and the putative LCC signal flows.
If we do take them as suggesting possible relationships between 
the functions of the two networks, 
we are led to at least a rough and tentative 
correspondence between
(a) NN layer R, and LCC layers 4 and 2/3; 
(b) NN layer Z, and LCC layer 6;
(c) NN layers g and T, and LCC layer 5;
(d) NN inputs $y$ to layers R and Z, and LCC sensory inputs 
to layers 4 and 6, 
respectively; 
(e) motor outputs $\tilde{u}$ and ${\rm out}_M$, with an 
efferent copy to 
NN layer Z
for prediction of the future plant state;
(f) the optimal estimate $\hat{y}$, and
${\rm out}_S$;
and (g) the g $\rightarrow$ R path of KC learning (Fig.~\ref{3ab}b), and observed
LCC connections from layer $5 \rightarrow 2/3$ (not shown in D1 and D2). 
   
We treat the g and T layers together since their role is limited to control, 
and since LCC layer 5 appears to 
be most prominent in motor control areas of cortex.  
We suggest that the Kalman NN may require one layer (R) in place of two 
(cf. LCC layers 4 and 2/3) because
Kalman estimation does not involve the 
learning of higher-level features (e.g., orientation selectivity in V1), and
we expect that more sophisticated 
(e.g., more strongly nonlinear and context-sensitive) NN prediction methods
may require an additional layer as in the LCC.

Note that in our NN, $\tilde{F}$ (used for prediction) and $\tilde{F}'$ 
(used for learning of control) connect the same pair of NN layers in
opposite directions, and are learned together during system identification.
This suggests that
a biological network performing Kalman-like prediction and control might use 
a corresponding pair of functional mappings that are (approximately) the 
transpose of one another.

For an example of a mapping between the KPC NN and the LCC that is
more detailed than I think is warranted in view of the caveats discussed
below, the interested reader may see Appendix D. 

\subsubsection{Differences and caveats}

We have compared one type of artificial NN -- one that performs
Kalman estimation and control, and system identification, without
simplifications or approximations (beyond that entailed by 
approximating an expectation value over an ensemble by 
a finite-sample average) -- with a putative and simplified 
biological LCC.  In order to be able to draw confident inferences from
any resemblances that emerge, it would be important to know whether 
the resemblances are robust across (a) several types of NN coding schemes, 
(b) a variety of relevant NN prediction and control methods, and 
(c) uncertainties regarding the biological network.  These are open questions.  
Caveats and limitations therefore include the following:

\begin{enumerate}
\item Nature of the computational task to be performed:
If the LCC performs estimation (including prediction) and control,
it surely performs it in a more general fashion than does KPC --
including the learning of higher-level features, and other nonlinear and 
context-dependent analyses (e.g., Bayesian inference and the use of
generative models) --
although the functions performed by the LCC might subsume KPC as
simple special cases.    
\item NN coding schemes and neuronal dynamics:
Neuronal dynamics are much more complex than the NN operations 
allowed here.  Despite this, it is commonly (and often fruitfully) assumed   
that reduced or simplified NN models can capture relevant dynamical features
of biological neuronal networks.
As an example, the node activity in 
a nonlinear (sigmoidal) version of the NN used here is often identified with
an average neuronal firing rate; and connection weights, with 
synaptic 
efficacies.\footnote{Average firing rates 
must be nonnegative and synaptic efficacies
cannot change sign.
To modify our linear NN 
to satisfy these constraints,
one could replace (a) each node by a rectifier plus two nodes
having activities $(v,0)$ if $v>0$ and $(0,-v)$ if $v<0$,
and (b) each connection by a direct path plus a path having an inhibitory internode.
These changes would not affect our results.
Regarding the use of a sigmoid nonlinearity, it is unnecessary for 
our KPC NN algorithm, and we have not found any way in which it enables an 
improvement over the use of linear nodes for performing (linear) KPC.
}
However, 
other types of NNs 
represent and process information in a variety of 
ways (Haykin, 1999; Hertz, Krogh, \& Palmer, 1991), 
using, e.g.,
population coding, 
sparse coding, 
coding via precise spike timing (Rieke \& Bialek, 1999), 
and
the related use of synchronous or phase-locked firing 
or oscillations for conveying information,
switching between functional modes, 
and/or more efficient learning. 
Detailed neuronal dynamics also affect the relative timing of excitatory
and inhibitory effects, 
the occurrence of
bursting vs. tonic firing modes, etc., all of which are 
absent from our simple NN. 

If a particular type of NN coding supports the basic operations used here --
matrix-times-vector multiplication, addition of vectors, and bilinear 
Hebbian learning --
then the derivation of the KPC algorithm and architecture 
can proceed as described, essentially
unchanged at the level of abstraction depicted in Figs.~\ref{3ab} and \ref{3c}
(the signal flows) and 
Fig.~\ref{4} (the block diagram discussed in Appendix B.2),
although the particular way
in which a vector is multiplied by a matrix will depend on the type of NN coding 
used. 
When the NN coding supports a quite different set of operations, however, 
it is an open question whether and how neural KPC may be 
implemented, and what the resulting architecture will be.
\item Question of uniqueness of our design: 
For our allowed set of NN operations, our exploration of the 
design constraints for 
performing general KPC suggests that the resulting signal flow and circuit 
are substantially determined (apart from small variations), 
but we cannot rule out the possibility of a 
significantly different design.
\item Experimental limitations: 
Knowledge of the detailed LCC connectivity is 
substantial, but not complete;
e.g., an inhibitory cell may provide output to many layers, and the 
extent and importance of
some of these connections are not clear.
Knowledge of the LCC signal flows and their sequencing is also quite
incomplete.
\end{enumerate}

\section{Conclusion}

The approach taken here has been to (a) pose a well-defined computational task 
-- Kalman prediction and control --
that is a prototype of the more general and powerful
prediction and control
processes that are likely to be important 
in cortex;
(b) select a simple typical set of allowed NN operations,
rather than invoking more complex NN dynamics specially tailored to the task;   
(c) see whether a NN algorithm can be devised that does not
compromise, change, or limit the computational task;
(d) see what constraints that
task imposes on the NN's circuitry and signal flows;
and (e) compare the resulting NN circuitry with that of the biological 
system of interest.

We have shown how optimal Kalman estimation and control,
and system identification, can be learned and 
executed by a neural network having 
simple computational elements. 
In progressing from the classical Kalman equations to a NN algorithm, 
we have find that the computational task appears to impose 
significant constraints on the 
resulting NN architecture, circuitry, and signal flows. 

When we compare the resulting architecture to that of 
recurrent neural circuits found
in brain, and to LCC architecture 
in particular, we find certain resemblances.
The LCC architecture has been suggested to perform core functions 
(as yet unknown) that underlie sensory and motor processing in general. 
It is plausible that such functions may include 
prediction, the estimation or inference of missing or noisy sensory data, 
and the goal-driven generation of 
control signals.  Thus the resemblances found between the KPC NN 
architecture and that of the `local cortical circuit' may not be
coincidental. 

However, before one can infer from such resemblances that the
LCC is likely to be performing a particular set of core functions, 
much more evidence is required.
The present work fits into a broader context of ongoing NN research by
many workers, 
in which (a) a range of biologically 
realistic neural coding strategies (e.g., population or spike coding)
is being studied, to see how simple neural computations can be 
carried out using these codes,
and (b) a range of computational tasks in prediction and control  
(e.g., Bayesian inference) is being explored, to see how recurrent NNs may 
implement such tasks.
From the experimental side, further elucidation of both the 
functional connectivity and signal flows in local cortical circuitry
is of course also required.
It will be interesting 
to see whether certain types of computational tasks are found to be 
associated with
specific architectural features in NNs and, if so, whether such mappings
have useful implications for understanding biological neural function.  

Such continuing interaction between theory and experiment may not only 
contribute to elucidating core aspects of cortical function, but
may also lead to insights into new methods for nonlinear 
estimation and control.

\section*{Acknowledgments}

I thank Drs.~Stan Goldin, Geoff Grinstein, and Roger Traub
for valuable comments on this and/or an earlier version of the manuscript.

\section*{Appendix A.  Mathematical details}

Here we prove that:
\begin{enumerate}
\item Eq.~\ref{eq:Zevol} and Eqs.~\ref{eq:KPm} are equivalent;  
\item $Z_t = E(\eta_t \eta'_t)$ and the $\eta$ evolution Eq.~\ref{eq:etaevol} 
(applied to each $\eta_t$ in the ensemble over which the expectation value
is defined) 
imply 
$Z_{t+1} = E(\eta_{t+1} \eta'_{t+1})$;
\item Eq.~\ref{eq:Tevol} and Eqs.~\ref{eq:LS} are equivalent;
and 
\item $T_t = E(w_t w'_t)$ and the $w$ evolution Eq.~\ref{eq:wevol} imply 
$T_{\tau-1} = E(w_{\tau-1} w'_{\tau-1})$.
\end{enumerate}
 
\subsection*{A.1  Neural Kalman estimation}

(1)  Eq.~\ref{eq:KPm} for $K_t$, and the definition $Z_t \equiv H P^-_t H' + R$, 
yield
$I-HK_t = I-H P^-_t H' Z^{-1}_t = I - (Z_t -R) Z^{-1}_t = RZ^{-1}_t$.
Thus Eq.~\ref{eq:KPm} yields
\begin{eqnarray}
Z_{t+1} & = & HF(I-K_t H) P^-_t F' H' + HQH' + R   \nonumber \\
& = & \tilde{F} (I-HK_t) H P^-_t H' \tilde{F}' H' + HQH' + R \nonumber  \\
& = & \tilde{F} R Z^{-1}_t (Z_t - R) \tilde{F}' H' + HQH' + R \nonumber  \\
& = & \tilde{F} (I-R Z^{-1}_t ) R \tilde{F}' H' + HQH' + R ~,
\end{eqnarray} 
which is Eq.~\ref{eq:Zevol}.

(2)  The plant and measurement equations for $x_{t+1}$ and $y_t$ yield
\begin{equation}
y_{t+1} = \tilde{F} Hx_t + \tilde{u}_t + Hm_t + n_{t+1} ~;
\end{equation}
thus
\begin{eqnarray}
\eta_{t+1} & = & -y_{t+1} + \tilde{F} y_t + \tilde{F} RZ^{-1}_t \eta_t + 
\tilde{u}_t 
\nonumber \\
& = & -Hm_t - n_{t+1} + \tilde{F} n_t + \tilde{F} R Z^{-1}_t \eta_t ~. 
\end{eqnarray}
Since (a) the noise terms $m_t$, $n_t$, and $n_{t+1}$ are mutually independent 
and have zero mean;
(b) $\eta_t$ depends on $n_t$ (through $y_t$) but not on $m_t$ or $n_{t+1}$;
(c) $R$ and $Z$ are symmetric matrices;
and (d) $E(m_t m'_t) = Q$, $E(n_t n'_t) = E(n_{t+1} n'_{t+1}) = R$, 
$E(\eta_t n'_t) = -E(n_t n'_t) = -R$, and $E(\eta_t \eta'_t) = Z_t$;
we obtain
\begin{eqnarray}
E(\eta_{t+1} \eta'_{t+1}) & = & H E(m_t m'_t) H' + 
E(n_{t+1} n'_{t+1}) + \tilde{F} E(n_t n'_t) \tilde{F}
+ \tilde{F} R Z^{-1}_t E(\eta_t n'_t) \tilde{F}' \nonumber \\
& & + \tilde{F} E(n_t \eta'_t) Z^{-1}_t R \tilde{F}'
+ \tilde{F} R Z^{-1}_t E(\eta_t \eta'_t) Z^{-1}_t R \tilde{F}'  \nonumber \\
& = & HQH' + R + \tilde{F} R \tilde{F}' - \tilde{F} R Z^{-1}_t R \tilde{F}'  
\nonumber \\
& = & \tilde{F} (I-RZ^{-1}_t) R \tilde{F}' + HQH' + R ~,
\end{eqnarray}
which equals $Z_{t+1}$ by Eq.~\ref{eq:Zevol}.

\subsection*{A.2  Neural Kalman control}

(3)  Since $T_{\tau} \equiv H'^+ S_{\tau} H^+ + \tilde{g}$,
$S_{\tau} = H' (T_{\tau}-\tilde{g}) H$.  Using Eqs.~\ref{eq:LS} and the 
definitions of 
$\tilde{g}$ and $\tilde{r}$,
\begin{eqnarray}
F' - L'_{\tau} B' & = & F' - F' H' (T_{\tau}-\tilde{g}) HB 
[ B'H' (T_{\tau}-\tilde{g}) HB + g ]^{-1} B'    \nonumber \\
& = & F' - F'H' (T_{\tau}-\tilde{g}) T^{-1}_{\tau} H'^+ \nonumber \\
& = & F' - F'H' (I-\tilde{g} T^{-1}_{\tau} ) H'^+ \nonumber \\
& = & H' \tilde{F}' \tilde{g} T^{-1}_{\tau} H'^+ ~.
\end{eqnarray}  
Thus
\begin{eqnarray}
S_{\tau -1} & = & (F'-L'_{\tau} B') S_{\tau} F + r \nonumber \\
& = & H' \tilde{F}' \tilde{g} T^{-1}_{\tau} H'^+ S_{\tau} H^+ H F + r \nonumber \\
& = & H' \tilde{F}' \tilde{g} (I - T^{-1}_{\tau} \tilde{g} ) \tilde{F} H + r ~;
\end{eqnarray} 
so
\begin{eqnarray}
T_{\tau -1} & = & H'^+ S_{\tau -1} H^+ + \tilde{g} \nonumber \\
& = & \tilde{F}' \tilde{g} (I-T^{-1}_{\tau} \tilde{g} ) \tilde{F} + 
\tilde{r} + \tilde{g} ~,
\end{eqnarray} 
which is Eq.~\ref{eq:Tevol}.

(4)  The internally generated noise terms $\nu^g_{\tau-1}$, $\nu^g_{\tau}$, and
$\nu^r_{\tau}$ are mutually independent and have zero mean.  The vector
$w_{\tau}$ depends on $\nu^g_{\tau}$ (by Eq.~\ref{eq:wevol}) but not on
$\nu^g_{\tau-1}$ or $\nu^r_{\tau}$, which are both generated only after 
$w_{\tau}$ has been computed, since the 
iterative calculation of $w$ proceeds in order of decreasing $\tau$.
Since $\tilde{g}$, $\tilde{r}$, and $T_{\tau}$ are symmetric matrices,
and 
$E[ \nu^g_{\tau} (\nu^g_{\tau})' ] =  
E[ \nu^g_{\tau-1} (\nu^g_{\tau-1})' ] = \tilde{g}$,
$E[ \nu^r_{\tau} (\nu^r_{\tau})' ] = \tilde{r}$,
$E(\nu^g_{\tau} w'_{\tau} ) = -\tilde{g}$,
and $E(w_{\tau} w'_{\tau} ) = T_{\tau}$, we obtain
\begin{eqnarray}
E( w_{\tau-1} w'_{\tau-1} ) & = & E[ \nu^g_{\tau-1} (\nu^g_{\tau-1})'] + 
E[ \nu^r_{\tau} (\nu^r_{\tau})']
+ \tilde{F}' E[ \nu^g_{\tau} (\nu^g_{\tau})'] \tilde{F} \nonumber \\ 
& & + \tilde{F}' \tilde{g} T^{-1}_{\tau} E(w_{\tau} w'_{\tau}) 
T^{-1}_{\tau} \tilde{g} \tilde{F} \nonumber \\
& & + \tilde{F}' E( \nu^g_{\tau} w'_{\tau} ) T^{-1}_{\tau} \tilde{g} \tilde{F}
+ \tilde{F}' \tilde{g} T^{-1}_{\tau} E[ w_{\tau} (\nu^g_{\tau})' ] \tilde{F} 
\nonumber \\
& = & \tilde{g} + \tilde{r} + \tilde{F}' \tilde{g} \tilde{F} 
- \tilde{F}' \tilde{g} T^{-1}_{\tau} \tilde{g} \tilde{F} ~,  
\end{eqnarray}  
which equals $T_{\tau-1}$ by Eq.~\ref{eq:Tevol}.

\section*{Appendix B.  Neural circuit and functional block diagram}

\subsection*{B.1  Mapping of signal flows onto the static circuit}

The following notes are intended to aid in the tracing of the signal flows 
of Fig.~\ref{3ab} through
the NN circuit wiring diagram of Fig.~\ref{3c}.

KF signal flow (without KC execution)
is shown by the solid lines in Fig.~\ref{3ab}a, 
which carry out the $\eta$ calculation of Eq.~\ref{eq:eta}.
The corresponding circuitry consists of
a subset of the
solid lines in Fig.~\ref{3c}.  KF signal flow
starts with $y$ input to the left circle of layer Z (denoted Z-left); 
result $\eta$ is multiplied by $Z^{-1}$ by passage
through the $Z$ or $Z^{-1}$ connections (see text, Methods 1 and 2)
to Z-right; result is conveyed to R-right, then is multiplied by $R$ 
by passing through the $R$ connections to R-left, where
$y$ is added; result $\hat{y}$ is multiplied by the $\tilde{F}$ connections 
from R-left 
to g-left; result $\tilde{F}\hat{y}$ is conveyed to Z-left, where the 
cycle repeats for the incremented value of $t$.  [An external control term
$\tilde{u}({\rm ext})$, if present, is added at Z-left (not shown).]    

KC execution 
(cf. Eq.~\ref{eq:u})
adds the following computation, shown by the dashed
lines in Fig.~\ref{3ab}a and a subset of the solid lines in Fig.~\ref{3c}:  
$\tilde{F}\hat{y}$ at g-left is
multiplied by $\tilde{g}$, passing to g-right; result is sent to T-right and
multiplied by $T^{-1}$, passing to T-left; here $\tilde{F}\hat{y}$ is
subtracted, via the direct link from g-left to T-left,
yielding $\tilde{u}$, which is sent as output and also (as an efferent copy) 
to Z-left, where it is added to the $\tilde{F}\hat{y}$ computed during KF flow.

Finally, KC learning
(cf. Eq.~\ref{eq:wevol}) is represented by
the dashed lines of Figs.~\ref{3ab}b and \ref{3c}, 
and the solid-line lateral connections of Fig.~\ref{3c}.
It proceeds from T-left (with subtractive
input $\nu^g$ yielding activity $w$), to T-right (being multiplied by $T^{-1}$),
thence to g-right; to g-left with multiplication by $\tilde{g}$;
result receives additive input $\nu^g$, is multiplied by $\tilde{F}'$ enroute 
to R-left;
and passes to T-left, repeating the cycle for the decremented value of $\tau$.

\subsection*{B.2  Block diagram of the composite neural algorithm}

\begin{figure}[htb]
\includegraphics[width=6in]{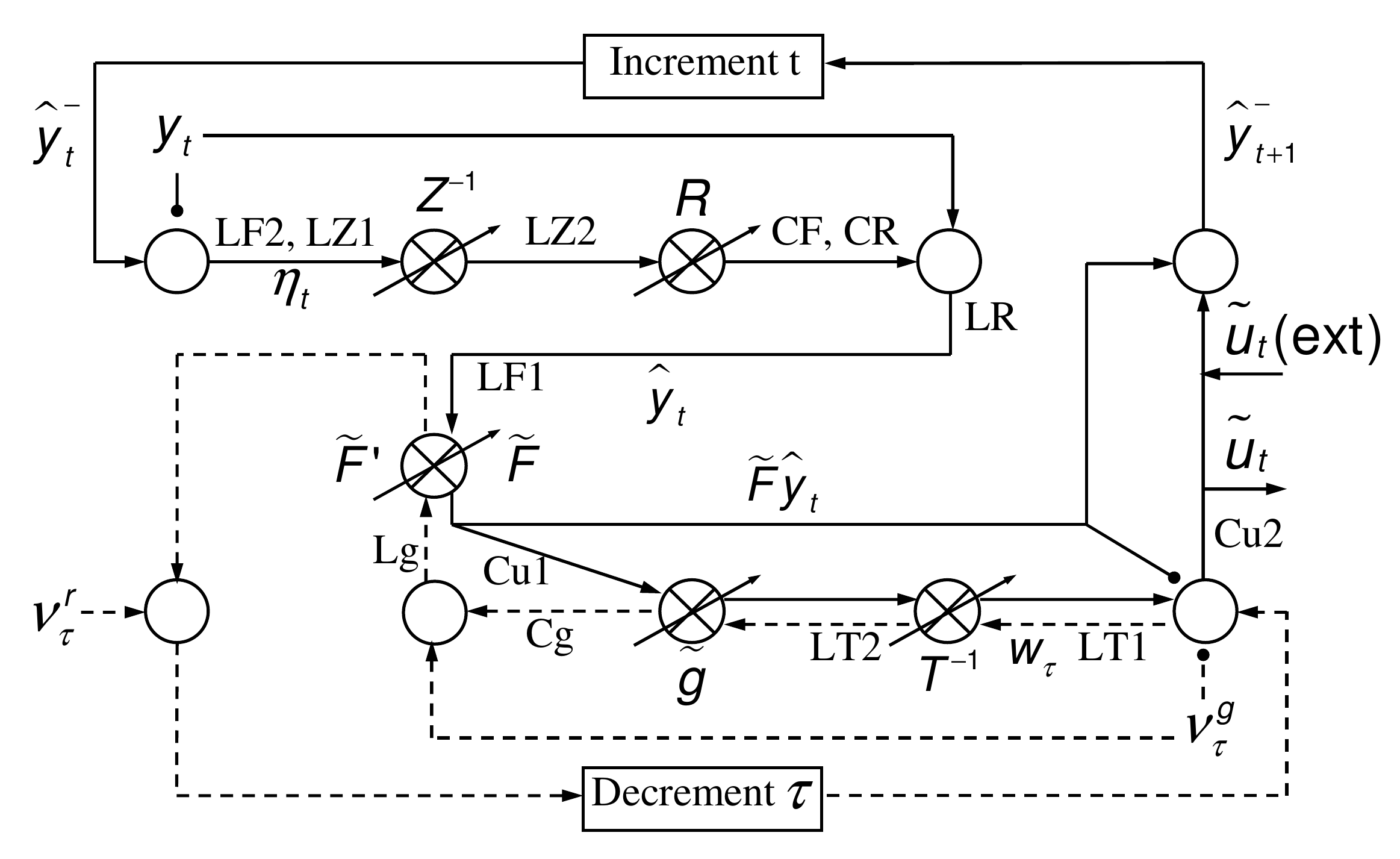}
\caption{ 
Block diagram of full NN algorithm.
Each $\bigcirc$ block combines inputs;
arrowhead inputs are added, filled-circle inputs are subtracted.
Each $\otimes$ block multiplies its input by the matrix 
indicated alongside the block. 
An arrow through a block denotes a matrix to be learned.
The solid path implements KF learning and execution, and KC execution;
the dashed path, KC learning.
Link symbols are as in Fig.~\ref{1a}.
}
\label{4}
\end{figure}

Figure \ref{4} depicts the complete KPC NN algorithm in block diagram form, showing
signal flows as proceeding through a set of functional blocks, but 
at a level of abstraction higher than that of a specifically NN implementation.  

The solid-line portion of Fig.~\ref{4} shows the signal flow that integrates 
Kalman estimation (KF execution and learning), system identification, and KC execution.  
Assume for now that line Cu1 and/or Cu2 is cut; i.e., no control signals 
are computed.
For the Kalman estimation (KF)
execution process, starting with the link (near upper left) labeled $\eta_t$,
the computation sequence is (cf. Eq.~\ref{eq:etaevol}):
$\eta_t \rightarrow Z^{-1} \eta_t \rightarrow RZ^{-1} \eta_t \rightarrow
\hat{y}_t = y_t + RZ^{-1} \eta_t \rightarrow \tilde{F} \hat{y}_t \rightarrow
\hat{y}^-_{t+1} = \tilde{u}_t({\rm ext}) + \tilde{F} \hat{y}_t \rightarrow
\eta_{t+1} = -y_{t+1} + \hat{y}^-_{t+1}$ [external driving term 
$\tilde{u}_t({\rm ext})$ is optional].
Using Method 1 of section 3.3 (Eq.~\ref{eq:Dlearn}),
$Z$ learning occurs at link LZ1; or, using Method 2 (Eq.~\ref{eq:Zinvlearn}),    
$Z^{-1}$ learning occurs at link LZ2.
$\tilde{F}$ learning (Eq.~\ref{eq:Ftil1learn}) uses $\hat{y}_{t-1}$ at link LF1 
(held from the previous time step $t-1$)
and $\eta_{t}$ at link LF2.
For initial learning of $\tilde{F}$ (before $Z$ is used; 
cf.~Eq.~\ref{eq:Ftil0learn}),
the circuit is cut at link CF, so that LF1 carries activity $y_{t-1}$ and LF2
carries $\epsilon_{t} = \tilde{F}_{t-1} y_{t-1} + \tilde{u}_{t-1} - y_t$.  
In the `offline sensor' mode for learning $R$ (cf. section 3.5.2), 
the circuit is cut at link CR, so that 
the following link
LR carries activity $y_t = n_t$.

Execution of control is shown in Fig.~\ref{4} (solid path) starting at link Cu1.
The computation sequence (cf.~Eq.~\ref{eq:u}) is:     
$\tilde{F} \hat{y}_t \rightarrow \tilde{g} \tilde{F} \hat{y}_t \rightarrow
T^{-1} \tilde{g} \tilde{F} \hat{y}_t \rightarrow
\tilde{u}_t = -\tilde{F}\hat{y}_t + T^{-1} \tilde{g} \tilde{F} \hat{y}_t 
= (-I + T^{-1} \tilde{g}) \tilde{F}\hat{y}$ at link Cu2.
Output $\tilde{u}_t$ is the control signal, and is also provided as efferent-copy 
feedback
to compute $\hat{y}^-_{t+1}$. 

The dashed portion of Fig.~\ref{4} shows the signal flows 
that implement learning of control; 
i.e., $w$ evolution 
and the learning of $T$ (or $T^{-1}$) and $\tilde{g}$.   
For $w$ evolution (Eq.~\ref{eq:wevol}), 
the computation sequence is (starting near lower right, at the 
link labeled $w_{\tau}$):  
$w_{\tau} \rightarrow T^{-1} w_{\tau} \rightarrow \tilde{g} T^{-1} w_{\tau} 
 \rightarrow \nu^g_{\tau} + \tilde{g} T^{-1} w_{\tau} 
 \rightarrow \tilde{F}' ( \nu^g_{\tau} + \tilde{g} T^{-1} w_{\tau} )
 \rightarrow \nu^r_{\tau} + \tilde{F}' ( \nu^g_{\tau} + \tilde{g} T^{-1} w_{\tau} )
 \rightarrow 
w_{\tau-1} = -\nu^g_{\tau-1} + \nu^r_{\tau} 
+ \tilde{F}' ( \nu^g_{\tau} + \tilde{g} T^{-1} w_{\tau} )$.

Either $T$ learning occurs at 
link LT1, or
$T^{-1}$ learning occurs at 
link LT2 (cf.~Eq.~\ref{eq:learnTinv} and just above it).
The $\tilde{F}'$   
connections will join the same nodes as $\tilde{F}$, but in the reverse direction 
(discussed below); thus they are learned along with $\tilde{F}$ (prior to being 
used for control) using the same Hebb
rule (Eq.~\ref{eq:Ftil0learn} or \ref{eq:Ftil1learn}).  
Matrix $\tilde{g}$ is learned (analogously to $R$ above) in a mode
of circuit operation in which the dashed link Cg is cut,  
so that the next link
Lg carries activity $\nu^g_{\tau}$, which is used for Hebbian learning of 
$\tilde{g} = E[\nu^g (\nu^g)']$.  (Cf.~section 4, paragraph starting
`Learning of $\tilde{g}$ \ldots '.)  

The circuit of Fig.~\ref{4} operates switchably in several distinct modes, each 
using a portion of the circuit with cuts (breaks in signal flow) as described.
At start-up: Learn $\tilde{F}$ and $\tilde{F}'$;  
this circuit mode has a cut at CF, with signal flow along solid path.
Learn $R$ using offline sensor operation (no external input); this mode uses
solid path with cut at CR.
Then, for each (increasing) $t$:
\begin{enumerate}
\item Perform KF learning \& execution (solid path).  If not also performing KC, 
the KF mode has a 
cut at Cu1 or Cu2.
If performing KC:
\begin{enumerate}
\item If KC matrix for this $t$ has been stored, retrieve it.
Otherwise:
learn $\tilde{g}$ (mode: dashed path, cut at Cg) if not already done; 
iterate $\tau$ from goal-completion time $N$ backward to $t$ (mode: dashed path, 
$t$ held 
constant while $\tau$ is decremented);
and optionally store intermediate KC matrices. 
\item KC execution:  Use KC matrix to calculate $\tilde{u}$ (mode: solid path), 
and provide efferent copy to KF.
\end{enumerate}
\item Optionally update $\tilde{F}$ and $\tilde{F}'$ during KF operation 
(mode: solid path, no cut at CF).
\item Optionally update $R$ (`offline sensor' mode: CR cut, solid path).       
\end{enumerate}

\section*{Appendix C.  Comments on a recent KF-inspired NN algorithm}

In a recent paper, Szirtes, P\'{o}czos, \& L\H{o}rincz (2005) 
(hereafter 
referred to as `SPL') observe:
`Connectionist representation of [Kalman filter-like] mechanisms
with Hebbian learning rules has not yet been derived.'
They state:
`The first problem of the classical [Kalman] solution 
is that 
covariance matrices of [the plant and measurement] noises are 
generally assumed to be known.  The second problem is that \textellipsis 
the algorithm requires the calculation 
of a matrix inversion, which is hard to interpret in neurobiological terms.
[Here] we derive an approximation of the Kalman gain, which eliminates these 
problems.'

In this section we show that the sequence of alterations made by SPL
to the KF equations
is not well-justified, and
that the SPL simulations, which show a reduction of prediction error with time, 
do not provide evidence 
that their algorithm is in fact computing an approximation of the optimal 
Kalman gain.

SPL denotes the `Kalman gain' matrix by $K^t$, which corresponds to our $F K_t$ 
(cf.~our Eq.~\ref{eq:xhat} and their Eq.~3).
At the outset, their stated goal is actually not to compute the Kalman-optimal $K^t$.
Instead, they consider only a family of matrices, which we will denote 
$\tilde{K}(\theta)$, 
parametrized by a vector $\theta$, for which the elements of $\tilde{K}(\theta)$ 
are 
$\tilde{K}_{ij}(\theta) \equiv K_{ij} \theta_i$ for each column $i$,
where $K$ is a given and fixed arbitrary matrix.
SPL's goal is to adaptively learn $\theta$ over time, using a NN algorithm,
so that its final learned value 
$\theta^*$ has the property that 
$\tilde{K}(\theta^*)$
minimizes the prediction error 
over the family of all possible $\tilde{K}(\theta)$.
Since $K$ is arbitrary, the parametrized family of matrices
may not 
include (or even contain a matrix that approximates) the actual optimal KF.
(Note that if $K^t$ is an $N \times N$ matrix, the set of all possible $K^t$ is 
being replaced
by a family parametrized by $\theta$, which has only $N$, not $N^2$, components.)

Given this goal, the first step in the SPL derivation is to minimize the prediction 
error
by performing stochastic gradient descent. 
This yields a pair of equations for $\theta^{t+1}_k$ and an auxiliary matrix
$W^{t+1}_{ik}$, in terms of values at time $t$:
\begin{equation}
\theta^{t+1}_k = \theta^t_k + \alpha \Sigma_{lj} K_{kl} H_{lj} W_{jk} \epsilon^t_k ~;
\label{eq:SGth}
\end{equation}
\begin{equation}
W^{t+1}_{ik} = \Sigma_j F_{ij} W^t_{jk} 
- \theta^t_i \Sigma_{lj} K_{il} H_{lj} W^t_{jk} + \delta_{ik} \epsilon^t_k ~.  
\label{eq:SGW}
\end{equation}
To obtain their model 'O1' (SPL Eqs.~5-6), SPL introduces a random vector $\xi$, 
which `can be regarded as sparse, internally generated noise,' in order `to provide 
a conventional neuronal equation.'  In SPL Eq.~6, the first two right-hand terms 
are accordingly equal to $\xi_k$ times the corresponding terms of the above 
Eq.~\ref{eq:SGW}.  This multiplication by $\xi_k$ is, however, not justified. 
If, for example, $\xi$ has zero mean
(actually the $\xi$ distribution is nowhere specified),
each of those two (now incorrect) terms will average to zero, 
and will thus make no contribution, on average, to $W^{t+1}_{ik}$.  Therefore 
model `O1' is not a valid approximation.

Several additional alterations are then made to generate a succession of 
SPL models called `O2' through `O5.' 

To obtain model `O2' (SPL Eqs.~7-8), SPL writes `to simplify the complexity of the 
iteration, we may suppose that the system is near optimal: $K \approx H^{-1}$.'  
Since the $K$ referred to here is the arbitrary and fixed $K$ that enters the above 
definition of  
$\tilde{K}(\theta)$, one is free to choose $K \approx H^{-1}$, or even 
$K = H^{-1}$, 
irrespective of whether `the system is near optimal.'  However, doing so
in no way ensures that the learned $\theta$ will yield a $\tilde{K}$ that is 
approximately Kalman-optimal.   
Certainly the optimal KF is not in general approximately equal to $H^{-1}$; 
it may be quite far from 
this value, depending upon $F$ and the relationship between 
the plant and measurement noise covariances.  This (as well as some of the 
following points) can be
most easily confirmed by considering the simple 1-d case, in which all matrices 
are scalar
quantities.    

To obtain model `O3' (SPL Eqs.~9-10), SPL states that because only diagonal 
elements $W_{kk}$ enter SPL Eq.~8 for the evolution of $\theta$, therefore 
`we may neglect the off-diagonal elements of matrix $W$ in Eq.~7' 
(the evolution equation for $W$).  However,
such neglect will introduce errors into the {\it diagonal} elements 
of $W$, and thereby into the evolution of $\theta$.     

To obtain model `O4' (SPL Eqs.~11-12), SPL writes that a `further simplification 
neglects the self-excitatory contribution, 
$F_{ii} W^t_{ii} \xi_i$.'  However, this term is not in general small compared 
with the remaining terms.

Finally, to obtain model `O5' (SPL Eqs.~13-14), SPL introduces the 
`stabilized form' of model `O4.'  This means that instead of setting $W^{t+1}$ 
equal to a specified function $f(W^t)$, as in model `O4,' SPL re-defines 
$W^{t+1}$ as    
$W^{t+1} = W^t + \gamma f(W^t)$, where $\gamma$ is a (presumably small) 
learning rate.  
At this step, it would have been more appropriate to define instead
$W^{t+1} = (1-\gamma) W^t + \gamma f(W^t)$, so that if $W^{t+1} = W^t$ in 
model `O4,' it would do so in model `O5' as well. 

Thus each of models `O1' through `O5' is obtained by making an alteration 
or simplification that is not a justified approximation to the previous model, 
nor to the original stochastic gradient form given by our Eqs.~\ref{eq:SGth} 
and \ref{eq:SGW} above.

SPL then presents simulations (SPL Fig.~1) that show a rapid decrease in prediction 
error  
$\parallel x - \hat{x} \parallel$ (their $\hat{x}$ is our $\hat{x}^-$), 
and in the related `reconstruction error', $\parallel y-H \hat{x} \parallel$, 
with time.
This is stated to be a `comparison of direct [i.e., classical solution] and 
approximated Kalman
filters.'
However, it can easily be seen 
that even if one 
combines
the prediction from the previous time step with the current measurement
by using a blending matrix ($K^t$ in SPL's Eq.~3) 
that is
{\it arbitrary and fixed},
the prediction error starting with an initial arbitrary 
guess -- prior to making any measurements on the system --
will at first rapidly (exponentially) decrease as more measurements are made.
(This is true
provided only that $K^t$ 
is such that 
the posterior estimate of $x^t$ tends to move toward, rather than away from, 
$H^{-1} y^t$;
i.e., provided that $K^t$ actually blends the new measurement with the 
prior prediction,
rather than repelling the predicted $H \hat{x}$ away from the new measurement.)    

SPL does not display the $\tilde{K}(\theta)$ that is learned using their algorithm,
so one cannot tell whether or how that matrix converges to an approximation of
the optimal Kalman filter.
The asymptotic prediction errors obtained using (a) the optimal KF,
(b) an arbitrary fixed KF, and (c) a KF that varies according to an arbitrary 
algorithm,
will, subject to the proviso above, 
typically differ by a modest factor as long as the plant and measurement SNRs 
differ by only 8 dB
as in SPL's simulation.  Thus the differences in asymptotic error are not 
visible on the scale 
of SPL's Fig.~1, in which initial errors are huge (simply because the initial 
prediction is a guess
made in the absence of prior measurements) compared with the asymptotic errors 
using any of 
these blending matrices.  

It is also worth noting that the learning rate 
$\alpha = 0.01$ is held constant in SPL's simulation (according to the SPL Fig.~1 
caption).  
However, a rate factor that is appropriate 
initially, when prediction errors are huge, must be increased 
substantially as prediction errors decrease, in 
order to ensure that learning continues. 
This raises the possibility that SPL learning may have stalled early in the 
simulation,
although one cannot tell without seeing the evolution of $\theta$ or 
$\tilde{K}(\theta)$
with time.  Even if $\alpha$ had been adaptively altered to avoid such stalling, 
however, the above discussion shows that the claim of approximately optimal 
KF learning  
using SPL's `O1' through `O5' algorithms 
is not warranted, and that the comparison of predicted errors vs.~time in 
SPL's Fig.~1
does not indicate that KF-like learning has occurred.

\section*{Appendix D.  An alternative `mapping' between the KPC NN and
the LCC}

To see what happens if we consider a mapping that is 
more detailed than
I think is warranted in view of the caveats in Section 5, 
we can modify Dgm.~D1 to obtain Dgm.~D3: 
\begin{center}
\begin{tabular}{cccccccccr}
R & $\rightarrow$ & g & $\rightarrow$ & T & $\rightarrow$ & Z 
& $\rightarrow$ R  \\
$\uparrow$  &  & $\downarrow$	& & $\downarrow$  & & $\uparrow$ & 	\\		
$y$ & & $(\tilde{F}\hat{y})$ & & $\tilde{u}$  &	& $y$	& {\bf [D3]} \\	
\end{tabular}
\end{center}
Additional direct g $\rightarrow$ Z and Z $\rightarrow$ g paths 
(Fig.~\ref{3ab}a) are omitted 
from D3 for notational 
simplicity.  KC learning adds paths T $\rightarrow$ g $\rightarrow$ R 
(Fig.~\ref{3ab}b).   

On this view, one might then posit 
a one-to-one correspondence between NN layers \{R, g, T, Z\} and LCC 
layers \{4, 2/3, 5, 6\}, respectively.
Then an exact match would imply the existence of LCC paths
$4 \rightarrow 2/3 \rightarrow 5 \rightarrow 6 \rightarrow 4$,
$5 \rightarrow 2/3$ (used for KC learning in the NN), and $2/3 \rightarrow 6$,
and with 
input from a `lower' cortical area (or thalamus)
to 4 and 6, output from 5 to a lower area, and 
output from 2/3 to the same or a higher area.
All of these paths fit within current understanding of 
LCC connectivity
(Callaway, 1998; Douglas \& Martin, 2004; Gilbert, 1983; Raizada \& Grossberg, 2001).
An exact match would, however, also imply LCC paths 
$6 \rightarrow 2/3 \rightarrow 4$.
The $6 \rightarrow 2/3$ path 
has been described (Hirsch et al., 1998), 
but in the context of V1 complex-cell 
interconnections.  
Also, NN layer g is  
used only for KC (and as a layer through which
$\tilde{F}\hat{y}$ passes (Fig.~\ref{3ab}a)), 
arguing against identifying layer g with LCC layer 2/3,
which is important in sensory processing.
I thus expect the less-detailed resemblance between Dgms.~D1 and D2 
to be more robust than that between 
D3 and D2, as 
more complex prediction and control tasks,
and different sets of allowed NN operations,
are studied in future work.

\newpage
\section*{References}

Becker, S. \& Hinton, G. E. (1992).
Self-organizing neural network that discovers surfaces in
random-dot stereograms.
{\it Nature, 355}, 161-163.

Callaway, E. M. (1998).
Local circuits in primary visual cortex of the macaque monkey.
{\it Annu. Rev. Neurosci., 21}, 47-74.

Douglas, R. J. \& Martin, K. A. C. (2004).
Neuronal circuits of the neocortex.
{\it Annu. Rev. Neurosci., 27}, 419-51.

George, D. \& Hawkins, J. (2005).
A hierarchical Bayesian model of invariant pattern
recognition in the visual cortex.
{\it Proc. 2005 IEEE Int. Joint Conf. Neural Networks, 3}, 1812-17. 

Gilbert, C. D. (1983).
Microcircuitry of the visual cortex.
{\it Annu. Rev. Neurosci., 6}, 217-247.

Grossberg, S. \& Williamson, J. R. (2001).
A neural model of how horizontal and interlaminar connections
of visual cortex develop into adult circuits that carry out
perceptual grouping and learning.
{\it Cerebral Cortex, 11}, 37-58.

Haykin, S. (1999)
{\it Neural Networks: A Comprehensive Foundation},
2nd ed. (Prentice Hall).

Haykin, S. (2001)
{\it Kalman Filtering and Neural Networks}
(Wiley-Interscience); especially
chap. 2, pp. 23-68 by Puskorius, G. V. \& Feldkamp, L. A.
Parameter-based Kalman filter training:  Theory and implementation.

Hertz, J., Krogh, A., \& Palmer, R. G. (1991)
{\it Introduction to the Theory of Neural Computation}
(Addison-Wesley).  

Hinton, G. E. \& Ghahramani, Z. (1997).
Generative models for discovering sparse distributed representations.
{\it Philos. Trans. R. Soc. London, B, 352}, 1177-90.

Hinton, G. E., Osindero, S. \& Teh, Y. (2006).
A fast learning algorithm for deep belief nets.
{\it Neural Computation, 18}, 1527-1554. 

Hirsch, J. A., Gallagher, C. A., Alonso, J.-M., \&  Martinez, L. M. (1998).
Ascending projections of simple and complex cells in layer 6 of 
the cat striate cortex.
{\it J. Neurosci., 18}, 8086-94. 

Kalman, R. E. (1960).
A new approach to linear filtering and prediction problems.
{\it Trans. ASME -- J. Basic Eng., 82}, 35-45.

Klimasauskas, C. C. \& Guiver, J. P. (2001).
Hybrid linear -- neural network process control.
{\it U.S. Patent \#6278962}. 

K\"{o}rding, K. P. \& Wolpert, D. M. (2004).
Bayesian integration in sensorimotor learning. 
{\it Nature, 427}, 244-47. 

Lee, T. S. \& Mumford, D. (2003).
Hierarchical Bayesian inference in the visual cortex.
{\it J. Opt. Soc. Am. A, 20}, 1434-48.

Lewicki, M. S. \& Sejnowski, T. J. (1997).
Bayesian unsupervised learning of higher order structure.
{\it Adv. Neural Info. Proc. Systems, 9}, 529-35.

Linsker, R. (1992).
Local synaptic learning rules suffice to maximize mutual
information in a linear network.
{\it Neural Computation, 4}, 691-702.

Linsker, R. (2005).
Improved local learning rule for information maximization and
related applications.
{\it Neural Networks, 18}, 261-265.

Mountcastle, V. B. (1998).
{\it Perceptual neuroscience: the cerebral cortex}  
(Harvard Univ. Press).

Murata, N., M\"{u}ller, K.-R., Ziehe, A., \& Amari,  S.-i. (1997).
Adaptive on-line learning in changing environments.
{\it Adv. Neural Info. Proc. Systems, 9}, 599-605.

Poggio, T. \& Bizzi, E. (2004).
Generalization in vision and motor control.
{\it Nature, 431}, 768-74.

Raizada, R. D. S. \& Grossberg, S. (2001).
Context-sensitive binding by the laminar circuits of
V1 and V2: A unified model of perceptual grouping,
attention, and orientation contrast. 
{\it Visual Cognition, 8}, 431-466; 
see refs. cited in Table 1.

Rao, R. P. N. (1999).
An optimal estimation approach to visual perception
and learning.
{\it Vision Res., 39}, 1963-89.

Rao, R. P. N. (2004).
Bayesian computation in recurrent neural circuits.
{\it Neural Computation, 16}, 1-38.

Rao, R. P. N. (2005).
Bayesian inference and attentional modulation in the 
visual cortex.
{\it NeuroReport, 16}, 1843-48.

Rao, R. P. N. \& Ballard, D. H. (1997).
Dynamic Model of Visual Recognition Predicts Neural
Response Properties in the Visual Cortex.
{\it Neural Computation, 9}, 721-63.

Rieke, F., Warland, D., de Ruyter van Steveninck, R.,
\& Bialek, W. (1999).
{\it Spikes: Exploring the Neural Code} 
(MIT Press).

Rivals, I. \& Personnaz, L. (1998).
A recursive algorithm based on the extended 
Kalman filter for the training of feedforward neural models.
{\it Neurocomputing, 20}, 279-94.

Singhal, S. \& Wu, L. (1989).
Training Multilayer Perceptrons with the Extended Kalman Algorithm. 
{\it Advances in Neural Information 
Processing Systems, 1},  
133-140. 

Szirtes, G., P\'{o}czos, B., \& L\H{o}rincz, A. (2005).
Neural Kalman-filter.
{\it Neurocomputing, 65-66}, 349-355.

Szita, I. \& L\H{o}rincz, A. (2004).
Kalman filter control embedded into the reinforcement
learning framework.
{\it Neural Computation, 16}, 491-499.

Todorov, E. (2005).
Stochastic optimal control and estimation methods adapted to
the noise characteristics of the sensorimotor system.
{\it Neural Computation, 17}, 1084-1108.

Tresp, V. (2001).
Method and arrangement for the neural modelling of a 
dynamic system with non-linear stochastic behavior.
{\it U. S. Patent \#6272480}.

Williams, R. J. (1992).
Training recurrent networks using the extended Kalman filter.
In: {\it Proceedings of the International Joint Conference on Neural Networks}, 
June, 1992, Baltimore, MD, Vol. IV, 
pp. 241-246.

Yu, A. J. \& Dayan, P. (2005).
Inference, attention, and decision in a Bayesian neural 
architecture.
{\it Adv. Neural Info. Proc. Systems, 17}, 1577-84.

Zemel, R. S., Huys, Q. J. M., Natarajan, R., \&  Dayan, P. (2005).
Probabilistic computation in spiking populations.
{\it Adv. Neural Info. Proc. Systems, 17}, 1609-16.

\end{document}